\documentclass{ieeeaccess}
\usepackage{cite}
\usepackage{amsmath, amssymb, amsfonts}
\usepackage[english]{babel}
\usepackage{amsthm}
\usepackage{textcomp}
\usepackage{lipsum} 
\usepackage[hidelinks]{hyperref}
\newtheorem{theorem}{Theorem}[section]
\newtheorem{lemma}[theorem]{Lemma}

\usepackage{algorithm}
\usepackage{array}
\usepackage{algpseudocode}
\usepackage{graphicx}
\usepackage{textcomp} 
\usepackage{xcolor}
\usepackage{soul}

\usepackage{physics}
\usepackage{multirow}
\usepackage{subcaption}
\usepackage{mathtools}
\usepackage{comment}

\DeclarePairedDelimiter\ceil{\lceil}{\rceil}
\DeclarePairedDelimiter\floor{\lfloor}{\rfloor}
\def\BibTeX{{\rm B\kern-.05em{\sc i\kern-.025em b}\kern-.08em
    T\kern-.1667em\lower.7ex\hbox{E}\kern-.125emX}}
\newcolumntype{M}[1]{>{\centering\arraybackslash}m{#1}}

\begin{document}
\history{Date of publication xxxx 00, 0000, date of current version xxxx 00, 0000.}
\doi{10.1109/ACCESS.2017.DOI}

\title{Improvement of Spiking Neural Network With Bit Planes And Color Models}
\author{
    \uppercase{
        Nhan T. Luu\authorrefmark{1, 3, 4} (Member, IEEE), 
        Duong T. Luu\authorrefmark{2}, 
        Pham Ngoc Nam\authorrefmark{3} (Member, IEEE), 
        and Truong Cong Thang\authorrefmark{4} (Senior Member, IEEE)
        }
    }
\address[1]{College of Information and Communication Technology, Can Tho University, Can Tho, Vietnam}
\address[2]{Center for Digital Transformation and Communication, Can Tho University, Can Tho, Vietnam}
\address[3]{College of Engineering and Computer Science, VinUniversity, Hanoi, Vietnam}
\address[4]{Department of Computer Science and Engineering, The University of Aizu, Aizuwakamatsu, Japan}
\tfootnote{ Email: luutn@ctu.edu.vn, luutd@ctu.edu.vn, nam.pn@vinuni.edu.vn, thang@u-aizu.ac.jp }

\markboth
{Nhan T. Luu \headeretal: Improvement of Spiking Neural Network With Bit Planes And Color Models}
{Nhan T. Luu \headeretal: Improvement of Spiking Neural Network With Bit Planes And Color Models}

\begin{abstract}
Spiking neural network (SNN) has emerged as a promising paradigm in computational neuroscience and artificial intelligence, offering advantages such as low energy consumption and small memory footprint. However, their practical adoption is constrained by several challenges, prominently among them being performance optimization. In this study, we present a novel approach to enhance the SNN for image classification through a new coding method that exploits bit plane representation. Also, we investigate the impacts of color models of the proposed coding process. Through extensive experimental validation, we demonstrate the effectiveness of our coding strategy in achieving performance gain across multiple datasets of grayscale images and color images. To the best of our knowledge, this is the first research that considers bit planes and color models in the context of SNN. By leveraging the unique characteristics of bit planes, we hope to unlock new potentials in SNNs performance, potentially paving the way for more effective SNNs models in future researches and applications.
\end{abstract}


\titlepgskip=-15pt

\maketitle

\section{Introduction}
\label{sec:introduction}
\PARstart{T}{he} field of artificial intelligence (AI) has experienced unprecedented advancements in recent decades\cite{wani2020advances}, largely driven by the developments of deep learning algorithms. Despite these advancements, traditional artificial neural networks (ANNs) often fall short in modeling the intricate and dynamic nature of biological neural systems\cite{nguyen2021review}. To bridge this gap, spiking neural networks (SNNs) have emerged as a promising alternative, leveraging the temporal dimension of neuronal activity to more closely emulate the behavior of biological brains.

SNNs utilize discrete spike events to transmit information between neurons, mirroring the asynchronous and event-driven communication observed in biological neurons\cite{auge2021survey}. This approach not only enhances the biological plausibility of neural models but also offers potential advantages in terms of computational efficiency and power consumption, particularly for hardware implementations\cite{blouw2019benchmarking}. The inherent sparsity and temporality in coding of SNNs provide a robust framework for developing low-power neuromorphic systems that can operate efficiently in real-time environments \cite{rajendran2019low}.

Despite their potential, SNNs pose significant challenges, particularly in training methodologies and network optimization. Traditional backpropagation techniques used in ANNs are not directly applicable to SNNs due to their discontinuous and non-differentiable nature\cite{nunes2022spiking}. Consequently, novel training algorithms, such as spike-timing-dependent plasticity (STDP) \cite{srinivasan2017spike,liu2021sstdp} and surrogate gradient methods \cite{neftci2019surrogate, eshraghian2023training, fang2021deep}, have been developed to address these challenges, enabling the effective training of SNNs.

Prior research in ANN has explored the application of bit planes (e.g. \cite{vorabbi2023input}) and color models \cite{gowda2019colornet,taipalmaa2020different} to enhance performance across various tasks, demonstrating significant improvements in areas such as image recognition and processing. Motivated by these advancements, we aim to investigate whether bit planes and color models can similarly improve the performance of SNNs. This led us to formulate our research questions: 

\begin{itemize}
  \item \textit{How can we utilize bit plane representation to enhance the performance of SNN?}
  \item \textit{How can we utilize different color models to enhance the performance of SNN?}
\end{itemize}

In our research, we observe that bit planes of an image contain pulse-like information, and can be applied to encode pixel values for SNNs to improve accuracy during surrogate gradient training. Through extensive experiments, it is shown that bit planes contain valuable information that can be utilized in the spike coding process as a supplementary data source for SNN. Also, bit plane coding proves to be reutilized across multiple color models. Furthermore, RGB color model, which is the default color model of many datasets, is not always the best one among the evaluated color models in our study. 

A preliminary version of this paper is published in \cite{luu2024}, where the benefit of using bit plane coding is proved by three grayscale datasets. In this extended version, the following new points have been added.
\begin{itemize}
  \item First, the proposed method was updated to support color images.
  \item Second, eight more datasets were included in the experiments.
  \item Third, impacts of color models were investigated.
  \item Fourth, discussion and analysis of the paper were significantly extended.
\end{itemize}

\begin{table}[htb!]
    \centering
    \caption{Table of abbreviation (in alphabetical order) used in this paper.}
    \resizebox{\linewidth}{!}{
    \begin{tabular}{|M{2cm}|p{5.5cm}|} 
    \hline
    \textbf{Abbreviation} & \textbf{Full Name} \\
    \hline
    ADD & Addition\\
    \hline
    ANN & Artificial neural network\\
    \hline
    CIE & International Commission on Illumination\\
    \hline
    IF & Integrate-and-fire\\
    \hline
    LIF & Leaky integrate-and-fire\\
    \hline
    ReLU & Rectified linear unit\\
    \hline
    SEW & Spike-element-wise\\
    \hline
    SNN & Spiking neural network\\
    \hline
    SNR & Signal-to-noise ratio\\
    \hline
    STDP & Spike-timing-dependent plasticity\\
    \hline
    \end{tabular}}
    \label{tab:abbre}
\end{table}

\begin{table}[htb!]
    \centering
    \caption{Table of algorithmic symbols used in this paper.}
    \resizebox{\linewidth}{!}{
    \begin{tabular}{|M{2cm}|p{5.5cm}|} 
    \hline
    \textbf{Symbol} & \textbf{Meaning} \\
    \hline
    $X$ & Input image tensors\\
    \hline
    $X_{max}$ & Integer value indicate the maximum value of input image tensor\\
    \hline
    $n_{bit}$ & Minimum number of bit planes required to encode input $X$ into bit planes\\
    \hline
    $log_{2}()$ & Binary logarithm function\\
    \hline
    $X_{shape}$ & Dimensions of X, include:
    \begin{itemize}
      \item $B$: number of images being processed.
      \item $C$: number of image channels.
      \item $H$: image height.
      \item $H$: image width.
    \end{itemize}\\
    \hline
    $concat()$ & Tensor concatenation function, concatenating a list of tensors resulted to a tensor contain spikes of shape ($n_{bit}$, $X_{shape}$)\\
    \hline
    $convertColor()$ & Tensor color model converting function, convert $X$ to a specified color model decided by value of $colormodel$\\
    \hline
    $colormodel$ & A string value specify the color model to convert the tensor to (i.e. "RGB", "CMY")\\
    \hline
    $mod$ & Modulo division\\
    \hline
    $L$ & A list-like \cite{abelson1996structure} data structure, used to temporarily store coding result\\
    \hline
    $\mathbb{Z}$ & The set of all integers\\
    \hline
    \end{tabular}}
    \label{tab:sym}
\end{table}

\section{Spiking Neural Network}
In this section we present an overview of SNN. The abbreviations and algorithmic symbols used throughout the paper are listed in Table \ref{tab:abbre} and Table \ref{tab:sym}. 

\subsection{Overview of SNNs}
SNNs can be considered a biologically inspired variant of ANNs that offer a different computational paradigm. Unlike traditional ANNs, which process information in a continuous manner using floating-point activations and backpropagation-based learning \cite{rumelhart1986learning}, SNNs rely on discrete events called spikes to encode and transmit information \cite{maass1997networks}. This event-driven nature of SNNs allows for energy-efficient computations, making them suitable for neuromorphic hardware implementations \cite{merolla2014million}. While the majority of mainstream ANNs architecture operate on floating point data representations, SNNs mimic the sparse, time-dependent processing found in biological neurons \cite{gerstner2002spiking}. Furthermore, SNNs employ temporal dynamics to learn and process temporal information naturally, leveraging mechanisms such as STDP for unsupervised learning \cite{lee2018training}. Despite their advantages, the training of SNNs remains challenging due to the non-differentiable nature of spike events, often necessitating specialized training methods like surrogate gradients \cite{neftci2019surrogate, fang2021deep, luu2025accuracy}, hybrid SNN training \cite{luu2025hybrid} or ANN-to-SNN conversion approaches \cite{rueckauer2017conversion}. Thus, SNNs present a compelling alternative to ANNs for energy-efficient and biologically plausible computations, while also introducing unique challenges in model design and training.

One of the earliest neuron models that is still widely applied in modern SNN architectures is the perfect integrate-and-fire (IF) model (also sometimes referred to as the non-leaky integrate-and-fire model), first studied by Louis Lapicque in 1907 \cite{abbott1999lapicque}. In this model, the neuron is characterized by its membrane potential \( V \): 
\begin{equation}
    C \frac{dV(t)}{dt} = I(t)
\end{equation}
which evolves over time in response to an input current \( I(t) \). This evolution is described by the time derivative of the capacitance law, \( Q = CV \), where \( C \) is the membrane capacitance. As an input current is applied, the membrane potential increases until it reaches a fixed threshold \( V_{\text{th}} \). At this point, the neuron generates a spike modeled as a delta function, and the membrane potential is reset to its resting value, after which the process repeats. In this model, the firing rate increases linearly and indefinitely with the increase in input current.

While previous efforts have successfully converted ANN neurons with ReLU activations to IF neurons\cite{li2021free, deng2021optimal, hu2021spiking}, challenges still remain in the conversion process. One of the key issues is the difficulty in precisely determining the appropriate thresholds in these conversion approaches. Despite the introduction of both data-driven and model-driven threshold determination methods\cite{diehl2015fast}, the converted SNN often experience unstable accuracy compared to their re-trained ANN counterparts. Additionally, IF neurons lack time-dependent memory. Once a neuron receives sub-threshold potentials, the membrane potential persists without decay until the neuron fires. This behavior is said to be inconsistent with biological neurons, which exhibit dynamic temporal properties \cite{zhang2019tdsnn}. 

Still, recent study have shown that SNN employed with IF neurons are more energy efficient than traditional ANN\cite{dampfhoffer2022snns}. Within certain SNN architectures, IF neurons have demonstrated superior performance compared to the supposedly better than leaky integrate-and-fire (LIF) neurons\cite{zhang2019tdsnn}, with the SEW-ResNet\cite{fang2021deep} architecture being one such example.

\begin{figure*}[htb!]
\centerline{\includegraphics[width=\textwidth]{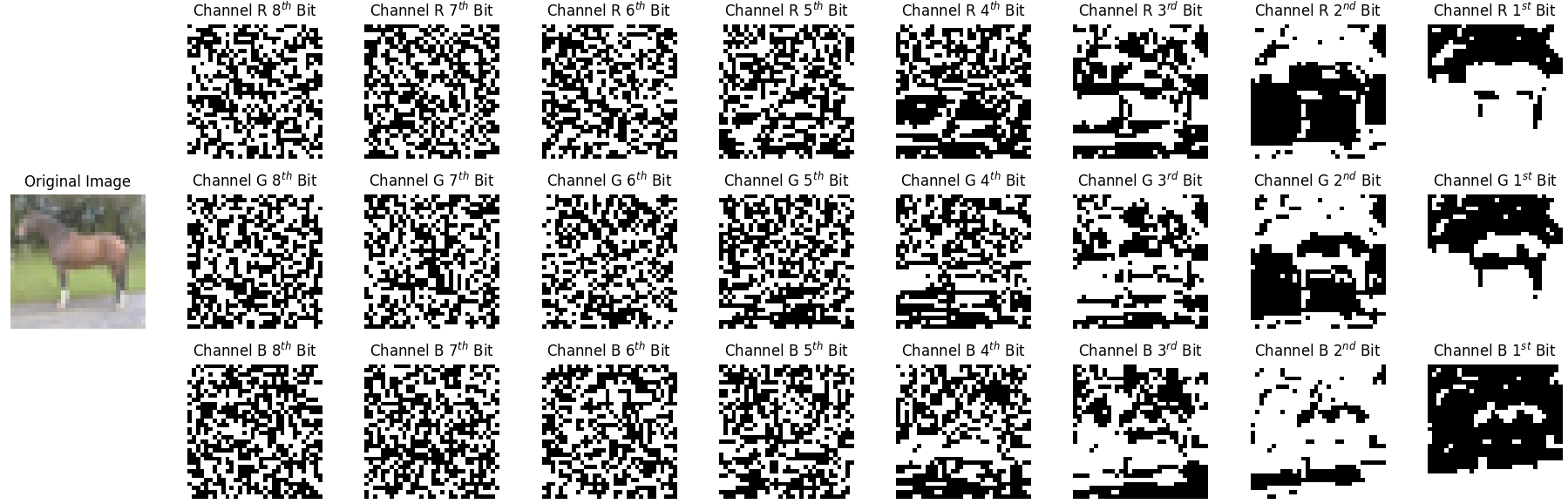}}
\caption{Visualization of the extracted bit planes from a sample image in the CIFAR-10 dataset \cite{alex2009learning}. Higher-order bit planes (e.g., most significant bits) encode more spatial and structural information, capturing the prominent features of the image. In contrast, lower-order bit planes (e.g., least significant bits) primarily contain finer details and noise, contributing less to the overall image structure.}
\label{bit_illustrate_fig}
\end{figure*}

\subsection{Deep Residual Learning in SNNs}\label{surrogate_grad_bg}

The success of ResNet\cite{he2016deep} in deep learning has motivated efforts to extend residual learning to SNN. Prior attempts, commonly referred to as Spiking ResNets\cite{hu2021spiking}, have largely replicated the architecture of ResNet\cite{he2016deep} by replacing ReLU activation functions with spiking neurons. However, this approach presents certain limitations:

\begin{itemize}
  \item \textit{Inability to achieve identity mapping across neuron models:} Residual learning is predicated on the concept of identity mapping, where layers can be trained to approximate an identity function. While this is straightforward in ResNets\cite{he2016deep} with ReLU activation, it is not always feasible in Spiking ResNets\cite{hu2021spiking}. Specifically, \cite{fang2021deep} claimed that for certain spiking neuron models, such as LIF neurons\cite{abbott1999lapicque} with learnable membrane time constants, ensuring that a spiking neuron fires in response to an input spike becomes challenging. This complicates the realization of identity mapping.
  \item \textit{Vanishing/exploding gradients:} Even in cases where identity mapping can be achieved, Spiking ResNets\cite{hu2021spiking} are prone to vanishing or exploding gradients during training. This occurs because the gradient propagation through multiple layers of spiking neurons may either diminish to zero or grow uncontrollably, complicating the training of deeper networks.
\end{itemize}

To address these issues, SEW ResNet\cite{fang2021deep} was introduced as a replacement design for residual blocks while utilizing IF neurons. The SEW block capitalizes on the binary nature of spikes by applying element-wise functions \( g \) to the output of the residual mapping \( A_l[t] \) and the input spike train \( S_l[t] \)\cite{fang2021deep}. This approach facilitates identity mapping, mitigates the vanishing/exploding gradient problem and achieves higher accuracy with fewer time steps compared to existing SNN models. 

In recent researches, using backpropagation methods to train SNNs \cite{huh2018gradient, eshraghian2023training, neftci2019surrogate, fang2021deep} has become a common practice in neuromorphic computing. In SNNs, spike signals are typically modeled using binary values (0s and 1s) from a Heaviside function $H(\theta, t)$, where $\theta$ is the membrane potential and $t$ is a threshold:

\begin{equation}
    H(\theta, t) = 
    \begin{cases}
      1 & \text{if $\theta - t \geq 0$}\\
      0 & \text{otherwise}
    \end{cases} 
\end{equation}

The derivative $\partial H$ of this function is given by:

\begin{equation}
    \frac{\partial H}{\partial \theta} = 
    \begin{cases}
      + \infty & \text{if $\theta - t = 0$}\\
      0 & \text{otherwise}
    \end{cases} 
\end{equation}

As shown, directly differentiating $H(\theta, t)$ results in unstable gradients. To address this, we can substitute the derivative of the Heaviside function $\partial H$ with a numerically stable gradient from a different activation function during backpropagation, such as the derivative $\partial \sigma$ of the Sigmoid function $\sigma(\theta, t)$:

\begin{equation}
    \begin{aligned}
        & \sigma(\theta, t) = \frac{1}{1+e^{\theta - t}}\\
        & \frac{\partial H}{\partial \theta} = \frac{\partial \sigma}{\partial \theta} = \frac{e^{\theta - t}}{(e^{\theta - t}+1)^2}\\
    \end{aligned}
    \label{eq:heaviside_sigmoid}
\end{equation}

This approach yields a more stable gradient during backpropagation, allowing the use of existing automatic differentiation frameworks to optimize SNNs effectively \cite{fang2021deep}.

Similar to other new types of neural networks \cite{ngo2023survey, nguyen2022evaluation, nguyen2022quantum}, the choice of input data representation may have strong impacts on output performance. Thus, in this paper we investigate a new way that employs bit plane coding for data representation in SNNs.

\section{Bit Planes and Color Models in Deep Learning}

This section presents an overview of bit planes and color models, along with their applications in deep learning algorithms.

\subsection{Bit planes in deep learning}
In the context of digital discrete signals such as images, a bit plane represents the collection of bits that occupy the same position within the binary representation of the signal's numerical values (as illustrated in Figure \ref{bit_illustrate_fig}). For instance, in a 8-bit data representation, there are 8 distinct bit planes. The $1^{st}$ bit plane consists of the most significant bits, while the $8^{th}$ bit plane comprises the least significant bits. The first bit plane provides the coarsest yet most fundamental approximation of the signal's values. As the bit plane index increases, its contribution to the overall representation diminishes. Consequently, incorporating additional bit planes progressively refines the signal's approximation.

Bit planes have shown extensive utility in deep learning applications, offering new ways to improve model efficiency and performance. In \cite{vorabbi2023input}, researchers used bit planes to encode the input layer in Binary Neural Network (BNN). They argued that bit planes retain crucial spatial information, enabling a reduction in the number of multiply-accumulate operations (MACs) without compromising the model's accuracy or robustness. This demonstrates the potential of bit planes in streamlining computations while maintaining competitive performance.

The use of bit plane slicing on RGB images for adversarial robustness was explored in \cite{liu2022defending}. By selecting the most significant bit planes, the authors developed a classification model that was better equipped to handle adversarial attacks. This approach capitalizes on the spatial information encoded in different bit planes, which helps the model resist manipulations that would otherwise degrade its performance.

In the domain of medical imaging, bit plane slicing has also been applied to improve accuracy in breast cancer recognition tasks\cite{chen2019breast}. Researchers found that lower-order bit planes are more susceptible to noise, and by removing them before inputting the data into a convolutional neural network (CNN), they achieved higher classification accuracy. This method highlights how bit plane analysis can be tailored to specific challenges, such as noise reduction in medical images, to enhance model effectiveness.

Although bit planes have been widely studied and applied in ANNs, their utilization in SNNs is still unexplored. As far as we are concerned, our research is first to explore application of bit planes on SNNs spike coding process.

\subsection{Color models in deep learning} \label{color_sect}

Color models are fundamental in computer graphics as they define how colors are represented, manipulated, and displayed for digital images. In computer graphics applications, understanding and effectively managing color models is crucial for tasks such as rendering realistic scenes, creating digital artwork, and designing user interfaces. Color models can be divided into \textit{device-oriented models}, \textit{user-oriented models} and \textit{device-independent models}  \cite{ibraheem2012understanding, plataniotis2001color}.

Device-oriented color models are influenced by the signals of the devices and the resulting colors are affected by the technology used for display. Widely utilized in various applications that require color consistency with hardware tools, examples include any hardware devices used for human visual perception, such as televisions and video systems. Popular color models for devices include RGB, CMY(K) and YCbCr. 

User-oriented color models are considered as a pathway between the observer and the device handling the color information. For our study, we mostly investigate cylindrically described models, including HSL and HSV. According to \cite{ibraheem2012understanding}, these models also inherit other special properties such as device dependence and cylindrical-coordinate representation.

Device-independent color models are essential for defining color signals without reliance on the specific characteristics of any particular device or application. These models play a crucial role in applications that involve color comparisons and the transmission of visual information across networks that connect various hardware platforms. We will mostly research color models defined by the International Commission on Illumination (CIE), including CIE XYZ and CIE LAB.

Color models are becoming increasingly important in deep learning algorithms, offering valuable information that enhances model performance in various tasks, such as image recognition\cite{kim2018efficient, gowda2019colornet}, image segmentation\cite{taipalmaa2020different} and adversarial attacks prevention\cite{chyad2025exploring}.

Despite extensive researches on the use of color models in ANN, their application in SNN remains unexplored and requires further investigation.

\begin{figure*}[htb!]
\centerline{\includegraphics[width=\textwidth]{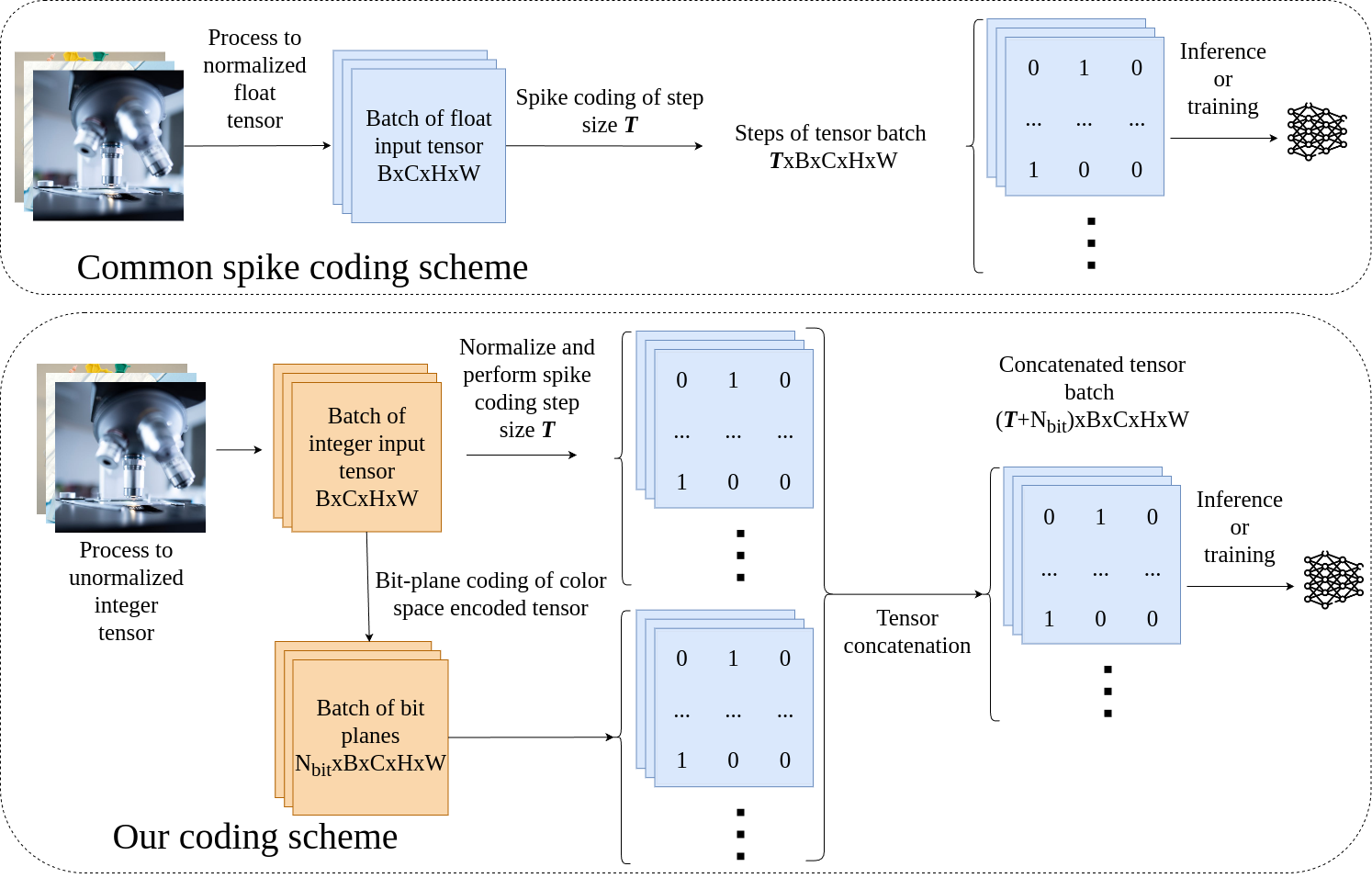}}
\caption{Our proposed coding method (lower) compare with traditional spike coding method (upper).}
\label{scheme_fig}
\end{figure*}

\section{Proposed Method}
\label{method}
In this section, we present in detail the design of our coding algorithm.


We hypothesize that the information contained in the bit planes of an image can be effectively used to encode inputs for SNN training. Bit planes decompose an image into binary levels, each representing a specific bit of the binary representation of pixel values. This decomposition can potentially preserve critical information of the original image while providing a format that might be more suitable for the spike-based processing in SNNs. The granularity of information available in bit planes offers a unique approach to encoding visual data, which could enhance the learning capabilities of SNNs by providing a richer, multi-level representation of the input.

To test this hypothesis, we conducted a preliminary experiment and analysis using the MNIST\cite{deng2012mnist}, which is a simple dataset containing grayscale images of hand-written numbers. We compared three approaches: traditional rate coding, bit plane coding only, and the combination of both rate coding and bit plane coding. The detailed settings of this experiment can be found in Section \ref{settings}. The results are provided in Table \ref{preliminary_table}.

\begin{table}[htb!]
    \centering
    \caption{Average validation accuracy comparison on MNIST\cite{deng2012mnist}}
    \resizebox{\linewidth}{!}{
    \begin{tabular}{|M{4.5cm}|M{3cm}|} 
    \hline
    \textbf{Coding method} & \textbf{Validation Accuracy} (\%)\\
    \hline
    Rate coding & 98.62\\
    \hline
    Bit plane coding only & 97.87\\ 
    \hline
    Combination of rate coding and bit plane coding & \textbf{98.93}\\ 
    \hline
    \end{tabular}}
    \label{preliminary_table}
\end{table}

We can see that the accuracy when using exclusively bit-plane-encoded inputs is lower compared to the baseline. However, a combination of rate coding and bit-plane coding resulted in an increase in performance. This outcome demonstrates that bit planes contain valuable information that SNN can learn from to improve their classification ability. This encourages us to carry out a further study and perform testing on other datasets.

As we can see in Figure \ref{scheme_fig}, the proposed coding scheme integrates a dual approach featuring both traditional spike coding and bit plane coding. Initially, the process commences with the conventional spike coding procedure applied to the color converted image tensor (defaulted to RGB if unspecified in our algorithm). The bit plane coding algorithm is then employed to transform an input image into binary planes. The final encoded tensors are then generated through tensor concatenation across the first dimension, merging the outputs from spike coding and bit plane coding. These encoded tensors are thereby prepared for utilization in inference tasks and/or training processes.

The process of converting a multi-channel image into its bit planes involves isolating each bit of the binary representation of the pixel values for each color channel. For an $n_{bit}$ image, let $I_{C}(x, y)$ denote the intensity of a pixel at position $(x, y)$ for a given color channel $C$, pixel intensity can be expressed in terms of its binary representation as:
\begin{equation}
    I_{C}(x, y) = \sum_{k=0}^{n_{bit}} B_{k, C}(x, y) \cdot 2^k
\end{equation}
where $B_{k, C}(x, y) \in \{0, 1\}$ is the $k$-th bit of the binary representation of $I(x, y)$, and $k$ represents the bit-plane index, with $k=0$ corresponding to the least significant bit and $k=n_{bit}$ corresponding to the most significant bit.

To isolate a specific bit-plane, the $k$-th bit of the pixel can be extracted using the following equation:
\begin{equation}
    B_{k, C}(x, y) = \left\lfloor \frac{I_C(x, y)}{2^k} \right\rfloor \mod 2
\end{equation}

\begin{algorithm}[t!]
\caption{Bit plane coding algorithm}
\label{bitplane_alg} 
\begin{algorithmic}
\Require $X_{max}$ is defined, $X_{max} > 0$ and $X \in \mathbb{Z}$
\Function{BitplaneEncode}{$X, X_{max}$, $colormodel=\text{"RGB"}$}
    \State $X = convertColor(X, colormodel)$ \Comment{Convert $X$ to specified color model}
    \State $ n_{bit} \gets \ceil*{\log_2(X_{max})}$ \Comment{Get number of bit required to encode image from highest possible value $X$}
    \State $ L \gets [\text{ }]$ \Comment{Empty list initialization}
    
    \For{each $X_{shape}.C$}
        \For{$i \gets 0, i < n_{bit}, i++$}
            \State $L.insert(X \mod{2})$ \Comment{Element-wise modulo of tensor, then store the bit plane to $L$}
            \State $X \gets \floor*{X/2}$ \Comment{Element-wise division of input image tensor and floor the result}
        \EndFor
    \EndFor
    
    \State \Return $concat(L)$ \Comment{Concatenate all stored sub tensors in $L$ to a single tensor}
\EndFunction
\end{algorithmic}
\end{algorithm}

To encode an image tensor batch $X$ into $n_{bit}$ bit planes in our training process based on established knowledge, we implement Algorithm \ref{bitplane_alg}. It is essential to know the maximum possible value $X_{max}$ of the tensor batch in order to determine $n_{bit}$. Since different color models have different $X_{max}$, this value must be predefined according to the selected color model. Once $X_{max}$ is defined, $n_{bit}$ can be calculate as:
\begin{equation}
    n_{bit} = \ceil{\log_2(X_{max})}
\end{equation}

Bit planes calculation of an unnormalized integer input tensor batch $X$ can be performed by element-wise division of $X$ by 2. The remainder from this division constitutes a bit plane of $X$, while the quotient is saved for the extraction of subsequent bit planes. This process is repeated until we obtain $n_{bit}$ bit planes. The resulting bit planes are then concatenated to form $n_{bit}$ binary tensor batch of $X_{shape}$. This format closely resembles that of rate coding and ensures determinism in the coding results without having to sample more than $n_{bit}$ times.

\section{Theoretical basis}
\label{theoretical}

In this part, we aim to establish a theoretical foundation for our proposed method of spike coding within surrogate gradient approaches.

As discussed in Section \ref{surrogate_grad_bg}, surrogate gradient methods attempt to adapt deep learning's gradient optimization techniques to SNNs, maintaining a close connection to traditional deep learning and convex optimization. 
A classical stochastic optimization problem associated with learning a target function $h$ can be formulated as\cite{shalev2017failures}:
\begin{equation}
    \min_{\mathbf{w}} F_h(\mathbf{w}) = \mathbb{E}_{\mathbf{x}} \left[ \ell(p_{\mathbf{w}}(\mathbf{x}), h(\mathbf{x})) \right],
\end{equation}
where $\ell$ is a loss function, $\mathbf{x}$ represents stochastic inputs (assumed to be vectors in Euclidean space), and $p_{\mathbf{w}}$ is a differentiable predictor parameterized by $\mathbf{w}$. It is assumed that $E_x[\left\|\frac{\partial}{\partial{\mathbf{w}}}p_{{\mathbf{w}}}\right\|^2] \leq G(\mathbf{w})^2$ for some scalar $G(\mathbf{w})$, and that $F$ is differentiable\cite{shalev2017failures, ghadimi2013stochastic}.

By reformulating our surrogate gradient optimization process as a traditional stochastic optimization problem, our objective can be defined as:
\begin{equation}
    \begin{aligned}
        \min_{\mathbf{w}} F_h(\mathbf{w}) &= \mathbb{E}_{\mathbf{x}} \left[ \ell(P_{\mathbf{w}}(\mathbf{x}), h(\mathbf{x})) \right], \\
        P_{\mathbf{w}}(\mathbf{x}) &= \mathbb{E}_{T} \left[ p_{\mathbf{w}}(e(\mathbf{x})) \right],
    \end{aligned}
\end{equation}
where the surrogate gradient model $P_{\mathbf{w}}(\mathbf{x})$ encodes the input $\mathbf{x}$ into a spike signal using an encoding function $e(\mathbf{x})$ and computes the expected output signal across the time dimension $T$ to generate the final stochastic prediction (basis for this statement can be found ar Appendix \ref{reusability}).

Previous studies in gradient-based optimization\cite{shalev2017failures, ghadimi2013stochastic} have demonstrated a positive correlation between a model's performance and the signal-to-noise ratio (SNR) of the stochastic gradient, defined by:
\begin{equation}
    \begin{aligned}
        & SNR(\delta(\mathbf{x})) = \frac{Sig(\delta(\mathbf{x}))}{Noi(\delta(\mathbf{x}))},\\
        & Sig(\delta(\mathbf{x})) = \left\| \mathbb{E}_{\mathbf{x}} \left[ \delta(\mathbf{x}) \right] \right\|^2\\
        & Noi(\delta(\mathbf{x})) = \mathbb{E}_{\mathbf{x}}\left\|\delta(\mathbf{x})-\mathbb{E}_{\mathbf{x}}\left[\delta(\mathbf{x})\right]\right\|^2\\
        & \delta(\mathbf{x}) = h(\mathbf{x}) \cdot g_\mathbf{w}(\mathbf{x})
    \end{aligned}
    \label{eq:snr}
\end{equation}
where $g_{{\mathbf{w}}}({\mathbf{x}}) =\frac{\partial}{\partial{\mathbf{w}}}p_{{\mathbf{w}}}({\mathbf{x}})$ is the Jacobian of model $p_{{\mathbf{w}}}({\mathbf{x}})$ with respect to model weight $\mathbf{w}$. Since this correlation does not depend on input gradient calculation, we can directly apply this in our research.
    
Upon performing signal concatenation using bit-plane-based signal $B$ of bit length $N_{bit}$ as in the proposed method, given certain condition satisfied, we can have:
\begin{equation}
    \begin{aligned}
        & \mathrm{sup}\left(SNR\left(\delta^{bit}(\mathbf{x})\right)\right) \geq  \mathrm{sup}\left(SNR\left(\delta(\mathbf{x})\right)\right),\\
        & \mathrm{sup}\left(SNR(\delta^{bit}(\mathbf{x}))\right) = \frac{\left\| \mathbb{E}_{\mathbf{x}} \left[ \delta^{bit}(\mathbf{x}) \right] \right\|^2_\infty}{\mathbb{E}_{\mathbf{x}}\left\|\delta^{bit}(\mathbf{x})-\mathbb{E}_{\mathbf{x}}\left[\delta^{bit}(\mathbf{x})\right]\right\|^2_\infty}, \\
        & \mathrm{sup}\left(SNR(\delta(\mathbf{x}))\right) = \frac{\left\| \mathbb{E}_{\mathbf{x}} \left[ \delta(\mathbf{x}) \right] \right\|^2_\infty}{\mathbb{E}_{\mathbf{x}}\left\|\delta(\mathbf{x})-\mathbb{E}_{\mathbf{x}}\left[\delta(\mathbf{x})\right]\right\|^2_\infty}\\
    \end{aligned}
    \label{eq:snr_ineq}
\end{equation}
and improving the chance of obtaining higher SNR (more background details and proof for Equation \ref{eq:snr_ineq} can be found at Appendix \ref{ineq_proof}). 

\begin{figure*}[t!]
\centerline{\includegraphics[width=\textwidth]{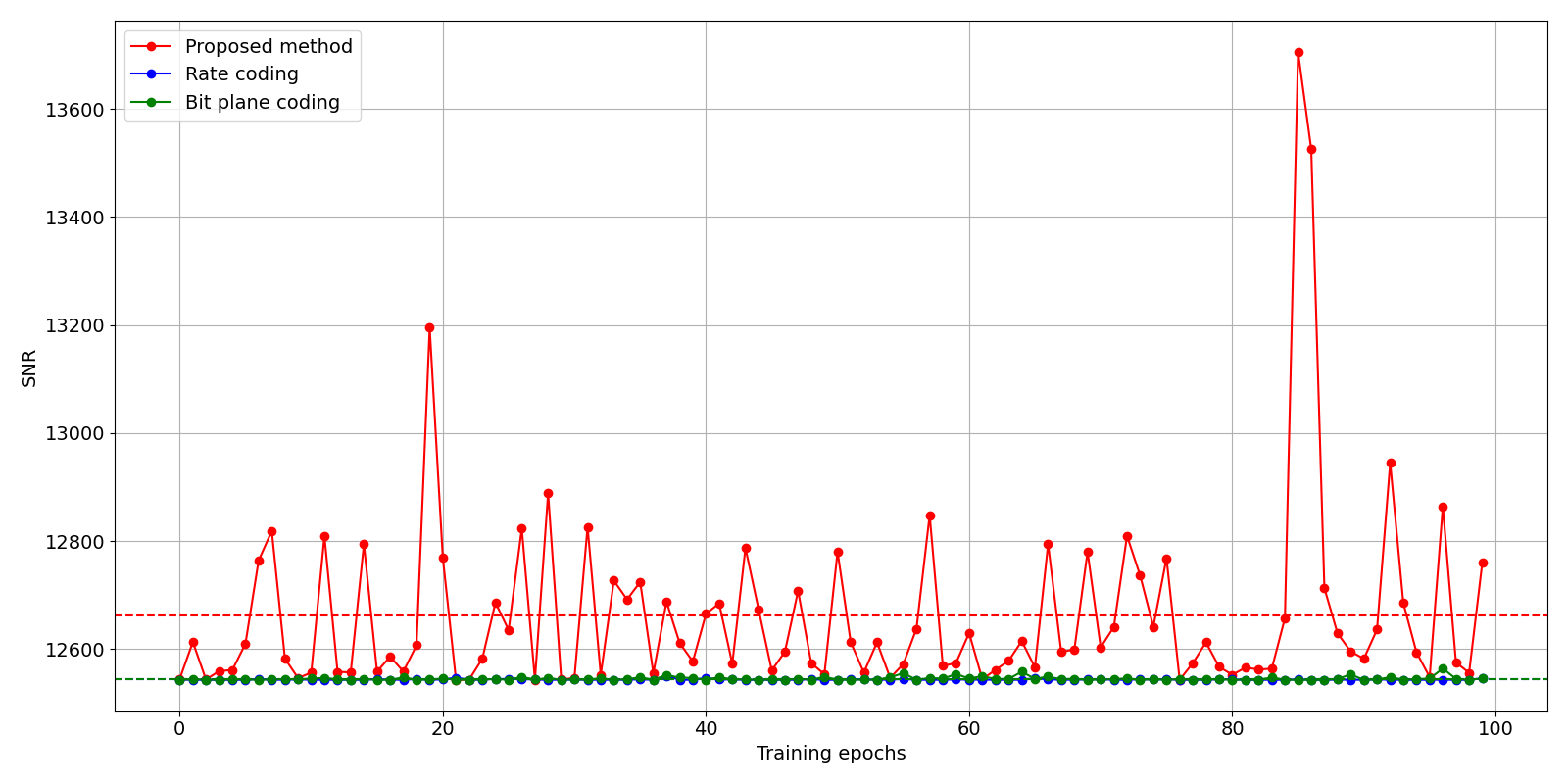}}
\caption{Comparing model gradient's SNR of the proposed method (red), baseline (blue) and bit plane coding (green) of MNIST dataset across the epochs. Dashed line indicate the mean gradient SNR (higher is better based on our assumption).}
\label{snr_fig}
\end{figure*}

To empirically validate our hypothesis, we measure the gradient SNR of the model across all training epochs on MNIST dataset (since to ensure that gradient does not saturate during training, \( p_{\mathbf{w}}({\mathbf{x}}) \) should serves as a reasonable approximation of the target function \( h({\mathbf{x}}) \)). The resulted SNR measurements across all training epochs are presented in Figure \ref{snr_fig}. The findings demonstrate that our proposed method achieves a higher mean SNR during training process, which aligns with our initial assumption. 

\color{black}
\section{Experiments}
\subsection{Experimental settings} \label{settings}
To investigate performance of SNN for images, we employed the popular tool SEW-ResNet18 with ADD element-wise function \cite{fang2021deep}. The model and experiment process were implemented using PyTorch \cite{paszke2019pytorch} and SpikingJelly\cite{doi:10.1126/sciadv.adi1480}. Color conversion algorithms are based on implementations from Kornia \cite{eriba2019kornia}. For the baseline method in the experiments, the \textit{Poisson-distributed rate coding} \cite{kim2018deep, heeger2000poisson} is employed with a step size $T = 10$ where spike $s_{T, i}$ for \textit{i}-th pixel at step $T$ with intensity $P_{T,i}$ can be interpreted as:
\begin{equation}
s_{T, i} = 
    \begin{cases}
      1 & \text{if } \mathcal{N}(0.0, 1.0) \leq P_{T, i} \\
      0 & \text{otherwise}
    \end{cases}
\end{equation}

To enable efficient training, surrogate gradient methods were applied to all IF neurons. We utilized the arctangent surrogate gradient function \( S(U) \)\cite{fang2021deep, eshraghian2023training}. Also, we adopted the standard cross-entropy loss function with a softmax activation, commonly adopted for classification tasks\cite{paszke2019pytorch, radford2021learning, luu2023blind}, computed using mean reduction across mini-batches.

Inspired by traditional deep learning approach in model fine-tunning \cite{radford2021learning, luu2023blind}, the optimization process is conducted with the Adam algorithm \cite{kingma2014adam}, using a learning rate of \( lr = 1 \times 10^{-3} \) and parameters \( \beta = (0.9, 0.999) \) across all evaluated datasets.

In the first part of our experiments, we only use grayscale image datasets to clearly see the effects of bit plane coding. In the second part, popular datasets of color images are used to see the practical effects of both bit plane coding and color models. We allocated 80\% of each dataset for training and the remaining 20\% with deterministic random seeding\footnote{Due to resource constraints, common validate tactics (such as utilizing k-fold for cross validation, etc.) might not be implementable on deep SNN training, where simulation of multiple time step is required and training cost are commonly higher than ANN on GPU devices. Enforcing deterministic random seed offer a good trade off for accuracy validation with certain level of bias in training and validating process.} for validation. All datasets were processed with a batch size of 16 images\footnote{While our main purpose for using small batch size is due to resource constraints, prior work have shown that model training with small batch size can still yield optimal solution given proper setting implemented\cite{masters2018revisiting}.} for 100 epochs on an NVIDIA RTX 3090 GPU.

\subsection{Experimental Results}
In this part, we will discuss results of our main experiments along with some interesting findings beside original hypotheses.

\subsubsection{The effect of bit plane coding on grayscale image datasets} \label{sub-results}

\begin{table}[hbt!]
    \centering
    \caption{Comparison of validation accuracy among coding methods. Best performances on each dataset are denoted in bold. Performance gain when compared with the baseline is highlighted in green.}
    \resizebox{\linewidth}{!}{
    \begin{tabular}{|M{2.5cm}|M{1cm}|M{1cm}|M{2cm}|}
    \hline
    \multirow{3}{*}{Dataset} & \multicolumn{3}{c|}{Accuracy of coding methods (\%)} \\
    \cline{2-4}
     & Baseline & Bit plane coding & Proposed method\\
    \hline
    MNIST\cite{deng2012mnist} & 98.62 & 97.87 & \textbf{98.93} (\textcolor{green}{+0.31})\\
    \hline
    KMNIST\cite{clanuwat2018deep} & 94.13 & 92.33 & \textbf{95.58} (\textcolor{green}{+1.45})\\
    \hline
    Fashion-MNIST\cite{xiao2017fashion} & 88.83 & 86.66 & \textbf{90.19} (\textcolor{green}{+1.36})\\
    \hline
    \textbf{Average} & 93.86 & 92.29 & \textbf{94.90} (\textcolor{green}{+1.04})\\
    \hline
    \end{tabular}}
    \label{tab:grayscale_acc_compare}
\end{table}

First, we aim to investigate whether the proposed approach yields any significant or insightful results without impacts of color information. For that, we employed three popular datasets of grayscale images, including MNIST\cite{deng2012mnist}, KMNIST\cite{clanuwat2018deep}, Fashion-MNIST\cite{xiao2017fashion}. The results of MNIST dataset were already shown in Table \ref{preliminary_table}. However, they are still included here to see the behaviors of the coding methods over different datasets. As grayscale images consist of only intensity values ranging between [0, 255], we apply our coding process with one channel. This process would produce 8 bit-plane channels that represent the image information in binary form.
In this part, we investigate three options,namely:
\begin{itemize}
  \item the baseline model (spike coding only),
  \item using bit plane coding only,
  \item the proposed model (spike coding combined with bit plane coding).
\end{itemize}

As demonstrated in Table \ref{tab:grayscale_acc_compare}, the combination of spike coding with bit-plane coding outperformed other coding methods in terms of validation accuracy. As mentioned, using only bit-plane coding resulted in a reduction in accuracy; however, the combination of spike coding with bit plane coding (i.e. the proposed method) provides gain in all datasets. For any method, the accuracy value is highest with the MNIST dataset and lowest with the Fashion-MNIST dataset. This could be because among these three datasets, MNIST is the simplest one while Fashion-MNIST is the most complex one. Interestingly, the gain of the proposed method compared to the baseline is increased from 0.31\% (with MNIST) to roughly 1.45\% (with KMNIST) and averaged at 1.04\%. That means the impact of bit plane coding is more prominent with more complex datasets.  

\subsubsection{Comparison of SNN variants}

\begin{table*}[htb!]
    \centering
    \caption{Comparison of validation accuracy of the baseline and the proposed method on different SNN variants. Performance gain when compared with the baseline is shown in green and best performing model on average is denoted in bold.}
    \resizebox{\textwidth}{!}{
    \begin{tabular}{|M{2.74cm}|*{5}{M{0.5cm}|M{1.54cm}|}}
    \hline
    \multirow{2}{*}{} & \multicolumn{10}{c|}{Accuracy of coding methods per dataset(\%)} \\
    \cline{2-11}
     &\multicolumn{2}{M{2.5cm}|}{MNIST}&\multicolumn{2}{M{2.5cm}|}{Fashion-MNIST}&\multicolumn{2}{M{2.5cm}|}{KMNIST}&\multicolumn{2}{M{2.5cm}|}{CIFAR10}&\multicolumn{2}{M{2.5cm}|}{\textbf{Average}}\\
    \hline
    SNN variant & Base-line & Proposed method & Base-line & Proposed method & Base-line & Proposed method & Base-line & Proposed method & Base-line & Proposed method\\
    \hline
    SEW-ResNet18 (ADD) & 98.62 & {98.93} (\textcolor{green}{+0.31}) & 94.13 &  {95.58} (\textcolor{green}{+1.45}) & 88.83 & 90.19 (\textcolor{green}{+1.36}) & 70.69 & {73.49} (\textcolor{green}{+2.80}) & 88.07 & \textbf{89.55} (\textcolor{green}{+1.48})\\
    \hline
    SEW-ResNet18 (AND) & 97.41 & 98.48 (\textcolor{green}{+1.07}) & 87.14 & 89.14 (\textcolor{green}{+2.00}) & 90.13 & 94.26 (\textcolor{green}{+4.13}) & 62.58 & 70.95 (\textcolor{green}{+8.37}) & 84.32 & 88.21 (\textcolor{green}{+3.89})\\
    \hline
    SEW-ResNet18 (IAND)  & 98.39 & 98.76 (\textcolor{green}{+0.38}) & 88.92 & 90.15 (\textcolor{green}{+1.23}) & 93.23 & 94.90 (\textcolor{green}{+1.67}) & 67.28 & 71.40 (\textcolor{green}{+4.12}) & 86.96 & 88.80 (\textcolor{green}{+1.84})\\
    \hline
    Spiking-ResNet18  &  98.25 & 98.74 (\textcolor{green}{+0.49})& 87.47 & 89.37 (\textcolor{green}{+1.90}) & 93.23 & {95.06} (\textcolor{green}{+1.83})& 46.15 & 61.79 (\textcolor{green}{+15.1}) & 81.28 & 86.24 (\textcolor{green}{+4.96})\\
    \hline
    Spiking-ResNet34   & 69.02 & 92.77 (\textcolor{green}{+23.8})& 69.59 & 79.62 (\textcolor{green}{+10.0}) & 37.45 & 72.23 (\textcolor{green}{+34.8}) & 19.28 & 32.29 (\textcolor{green}{+13.0}) & 48.84 & 69.23 (\textcolor{green}{+20.4})\\
    \hline
    Spiking-ResNet50   & 13.45 & 46.22 (\textcolor{green}{+32.8})& 12.80 & 57.20 (\textcolor{green}{+44.4}) & 10.44 & 23.87 (\textcolor{green}{+13.4}) & 17.94 & 27.78 (\textcolor{green}{+9.84}) & 13.66 & 38.77 (\textcolor{green}{+25.1})\\
    \hline
    \end{tabular}}
    \label{tab:opt_var_compare}
\end{table*}

In addition, we also evaluated our spike coding scheme in different SNN variants. We employ only both grayscale and color image datasets in this part, including MNIST, KMNIST, Fashion-MNIST and  CIFAR10. We benchmarked against multiple variants of Spiking ResNet\cite{hu2021spiking} with LIF surrogate neurons, which have demonstrated competitive performance in prior work, along with AND and IAND variants of SEW-ResNet18\cite{fang2021deep}.

In shown in Table \ref{tab:opt_var_compare}, the results reveal that in all of the tested variants, the proposed method consistently achieves a higher accuracy than the baseline. The performance gains range from 0.31\% (in the case of SEW-ResNet18 (ADD) on MNIST dataset) to 44.40\% (in the case of Spiking-ResNet50 on Fashion-MNIST). Also, Spiking-ResNet variants tend to underperform when compared with SEW-ResNet18 variants when benchmarking under spike-based backpropagation (which had been priorly noted by \cite{fang2021deep} due to gradient instability in previous works and Section \ref{surrogate_grad_bg}). The average accuracy values across all datasets show that SEW-ResNet18 (ADD) is the best variant. Fang et al. \cite{fang2021deep} also demonstrated that this variant usually achieves superior performance. Therefore, in subsequent experiments, we exclusively utilize this variant for benchmarking.

\subsubsection{The effect of bit plane coding on color images} \label{color-results}
\begin{table}[htb!]
    \centering
    \caption{Average validation accuracy on various computer vision dataset of our proposed method. Best performances are denoted in bold. Performance gain when compare with the baseline is highlighted in green.}
    \resizebox{\linewidth}{!}{
    \begin{tabular}{|M{2cm}|*{2}{M{1cm}|}M{1.7cm}|} 
    \hline
     \multirow{3}{*}{Dataset} & \multicolumn{3}{c|}{Accuracy of coding methods (\%)} \\
    \cline{2-4}
     & Baseline & Bit plane coding & Proposed method\\
    \hline
    CIFAR10\cite{alex2009learning} & 70.69 & 37.28 & \textbf{73.49} (\textcolor{green}{+2.80})\\
    \hline
    CIFAR100\cite{alex2009learning}  & 38.57 & 11.28 & \textbf{42.15} (\textcolor{green}{+3.58})\\
    \hline
    Caltech101\cite{li_andreeto_ranzato_perona_2022}  & 61.67 & 43.28 & \textbf{64.55} (\textcolor{green}{+2.88})\\
    \hline
    Caltech256\cite{griffin_holub_perona_2022}  & 30.04 & 21.75 & \textbf{41.51} (\textcolor{green}{+11.47})\\
    \hline
    EuroSAT\cite{helber2019eurosat}  & 84.39 & 79.87 & \textbf{88.48} (\textcolor{green}{+4.09})\\
    \hline
    Imagenette\cite{imagenette}  & 72.99 & 48.20 & \textbf{78.68} (\textcolor{green}{+5.69})\\
    \hline
    Food101\cite{food101}  & 19.62 & 6.29 & \textbf{36.95} (\textcolor{green}{+17.33})\\
    \hline
    \textbf{Average}  & 53.99 & 35.42 & \textbf{60.83} (\textcolor{green}{+6.84})\\
    \hline
    \end{tabular}}
    \label{tab:rgb_acc_compare}
\end{table}
\color{black}
For our experiment, we conducted tests on various color-image classification datasets, including CIFAR10\cite{alex2009learning}, CIFAR100\cite{alex2009learning}, EuroSAT\cite{helber2019eurosat}, Caltech101\cite{li_andreeto_ranzato_perona_2022}, Caltech256\cite{griffin_holub_perona_2022}, Imagenette\cite{imagenette} (which is a subset of ImageNet\cite{deng2009imagenet}) and Food101\cite{food101}. The default color model of these datasets is RGB. Due to computing resource constraints of the simulation tool, large images are downscaled to 224x224 pixels. So, among these datasets, CIFAR10\cite{alex2009learning}, CIFAR100\cite{alex2009learning} and EuroSAT\cite{helber2019eurosat} are trained with original input size while images of all other dataset are resized to 224x224 pixels.

The accuracy results of different cases are shown in Table \ref{tab:rgb_acc_compare}. We can see that, the accuracy of the proposed method is always higher than that of the baseline. The gain ranges from the lowest of 2.80\% (with CIFAR100) up to 17.33\% (with Food101 dataset), depending on the employed dataset. This suggests that the benefit of the proposed method when applying to color images is more significant than when applying to grayscale images.

\begin{table*}[htbp]
    \centering
    \caption{Validation accuracy comparison of proposed method among color models.}
    \resizebox{\linewidth}{!}{
    \begin{tabular}{|M{2cm}*{7}{|M{1.5cm}}|}
    \hline
    \multirow{3}{*}{Dataset} & \multicolumn{7}{c|}{Accuracy of coding methods (\%)} \\
    \cline{2-8}
     & With RGB encoded (Ours) & With CMY encoded (Ours) & With YCbCr encoded (Ours) & With HSL encoded (Ours) & With HSV encoded (Ours) & With CIE XYZ encoded (Ours) & With CIE LAB encoded (Ours)\\
    \hline
    CIFAR10\cite{alex2009learning} & 73.49 & 73.73 & 73.64 & 73.39 & 73.06 & 73.63 & \textbf{74.06} \\
    \hline
    CIFAR100\cite{alex2009learning} & \textbf{42.15} & 41.31 & 41.62 & 41.94 & 41.91 & 41.73 & 42.03 \\
    \hline
    Caltech101\cite{li_andreeto_ranzato_perona_2022} & 64.55 & 65.65 & 65.53 & 66.46 & \textbf{68.41} & 65.30 & 66.22\\
    \hline
    Caltech256\cite{griffin_holub_perona_2022} & \textbf{41.51} & 40.65 & 40.84 & 40.55 & 39.94 & 40.50 & 40.79\\
    \hline
    EuroSAT\cite{helber2019eurosat} & 88.48 & 86.61 & 88.11 & 88.74 & 88.81 & \textbf{88.87} & 87.48 \\
    \hline
    Imagenette\cite{imagenette} & 78.68 & 77.35 & \textbf{78.78} & 78.55 & 78.50 & 77.35 & 78.70\\
    \hline
    Food101\cite{food101} & 36.95 & 37.71 & 36.98 & \textbf{38.03} & 36.41 & 37.10 & 35.87\\
    \hline
    \textbf{Average} & 60.83 & 60.43 & 60.79 & \textbf{61.09} & 61.01 & 60.64 & 60.74\\
    \hline    
    \end{tabular}}
    \label{tab:rate_acc_compare}
\end{table*}

\begin{figure*}[t!]
    \centerline{\includegraphics[width=\textwidth]{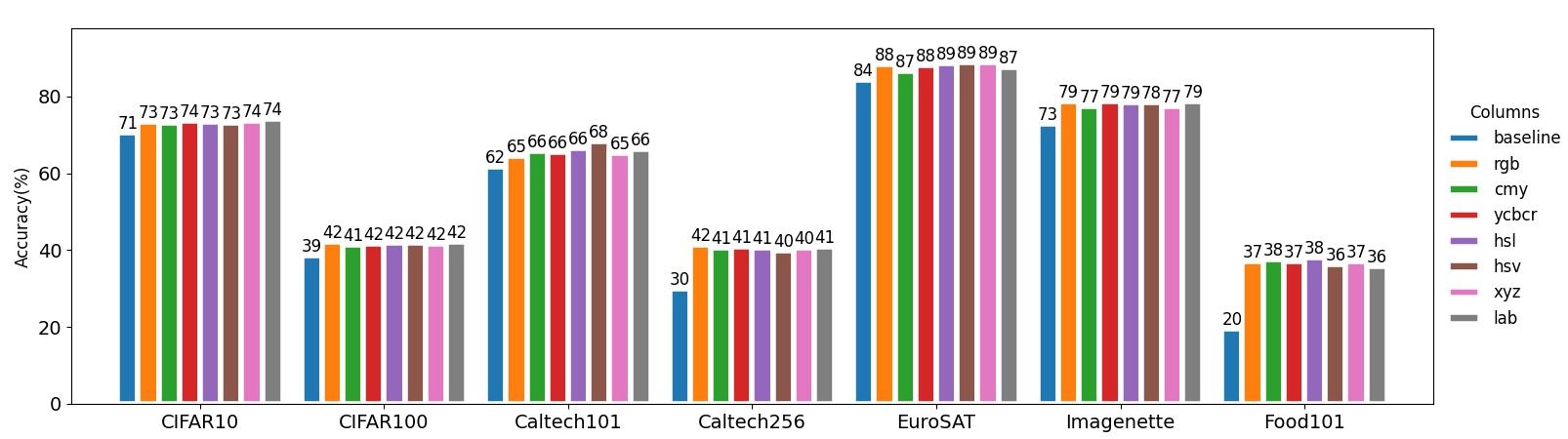}}
     \caption{Validation accuracy visualization of proposed method among color models.}
    \label{fig:rate_acc_compare}
\end{figure*}

\subsubsection{Impacts of color models} \label{main-results}
In this part, we investigate the effect of color models on the performance of proposed method. The results are provided in Table \ref{tab:rate_acc_compare} and also visualized in Fig. \ref{fig:rate_acc_compare}. In this figure, the baseline model is also included for the sake of comparison. From Fig. \ref{fig:rate_acc_compare}, it can be seen that our proposed method using different color models is consistently better than the baseline. Also, in general, the variations in performances of different color models are small.

From Table \ref{tab:rate_acc_compare}, the results show that RGB is not always the best color model. Among the eight datasets, RGB provides the best performance in only 2 cases (namely CIFAR100 and Caltech256). CMY does not achieve any best results, whereas each of the other five color models achieves the best performance once. Especially, with Caltech101 dataset, the accuracy of HSV color model is 68.41\%, which is considerably higher than that of RGB color model (64.55\%).

When averaging the accuracy values across all datasets, the HSL color model slightly outperforms the others for both coding method, as indicated in the last row of Table \ref{tab:rate_acc_compare}. The CMY color model, which does not achieve any best results, also has the lowest average performance. Further analysis reveals a distinctive characteristic of the HSL model that may contribute to its relatively stable performance. HSL encodes the hue component using a 0–359 degree representation, which results in a higher number of spike counts. Consequently, the maximum number of bits required for representation increases to $n_{bit} = 9$. As theoretically formulated in Section \ref{theoretical}, this greater bit depth might facilitates a higher SNR ratio that enables better performance. In contrast, the CMY model operates within a more limited range of $[0, 100]$ and thus needs only $n_{bit} = 7$ bits to represent the color values. Based on the same logic of Section \ref{theoretical}, this smaller number of bits reduces the gradient magnitude along with signal term for model SNR and thus results in its lower performance.

To rigorously quantify the computational overhead introduced by our method with different color models, we measure the running times of an image batch on forward propagation for both baseline and the proposed method. As shown in Table \ref{tab:perf_compare}, the increases in running times depend on datasets and color models, ranging from 20.52\% (with CIFAR-10 dataset and CMY color model) to 105.01\% (with CIFAR-10 dataset and HSL color model). Averaging over all datasets, the increases are from 48.57\% (with CMY color model) to 85.35\% (with HSL color model).

Figure \ref{fig:scatter_perf_avg} shows the relationship between the average gain in  accuracy and the average increase in processing time for different color models. We can see that any gain in performance requires some increase in processing time. Interestingly, while having the smallest performance gain, the CMY color model also has the lowest processing time increase. In contrast, the HSL color model has the greatest gain as well as the highest increase in processing time.

\subsection{Discussion}

Based on these observations, the key findings of our analysis are summarized as follows:

\begin{itemize}
    \item The proposed method yields performance improvements for both grayscale and color images by increasing gradient magnitude and improve model SNR during training process. 
    \item The gain of the proposed method depends on the employed dataset, ranging from 0.31\% to 1.45\% in small grayscale image dataset and 2.80\% to 17.33\% for practical colored image dataset with RGB-based variants.
    \item Proposed method can be generalized for different SNN variants.
    \item Though RGB is the default color model in existing datasets, it is not always the best one, providing highest results in only two datasets.
    \item Based on both theoretical foundation and in terms of average accuracy value over all datasets, HSL is the best color model and CMY is the worst one. Yet HSL provides the best performance in only one dataset.
\end{itemize}

\begin{table*}[htb!]
    \centering
    \caption{Comparison of model total forward and backward propagation time between baseline and proposed method using different color models. The values are average training times per epoch per images batch of 16}.
    \resizebox{\linewidth}{!}{
    \begin{tabular}{|M{2.4cm}|*{8}{M{1.4cm}|}} 
    \hline
     \multirow{3}{*}{Model Variants} & \multicolumn{8}{c|}{Processing time by dataset (ms)} \\
    \cline{2-9}
     & CIFAR-10 & CIFAR-100 & Caltech101 & Caltech256 & EuroSAT & Imagenette & Food101 & \textbf{Average}\\
    \hline
    Baseline & 41.67 & 43.55 & 141.25 & 146.92 & 66.05 & 135.71 & 140.66 & 102.26\\
    \hline
    Proposed (RGB) & 63.66 \textcolor{red}{(+52.77\%)} & 65.39 \textcolor{red}{(+50.15\%)}& 235.13 \textcolor{red}{(+66.46\%)} & 240.41 \textcolor{red}{(+63.63\%)} & 116.77 \textcolor{red}{(+76.79\%)} & 225.83 \textcolor{red}{(+67.07\%)} & 234.74 \textcolor{red}{(+66.89\%)}& 168.85 \textcolor{red}{(+65.12\%)}\\
    \hline
    Proposed (CMY) & 50.22 \textcolor{red}{(+20.52\%)}& 55.83 \textcolor{red}{(+28.20\%)}& 210.56 \textcolor{red}{(+49.07\%)}& 215.63 \textcolor{red}{(+46.77\%)} & 102.31 \textcolor{red}{(+54.90\%)} & 217.32 \textcolor{red}{(+60.78\%)} & 211.64 \textcolor{red}{(+50.46\%)}& 151.93 \textcolor{red}{(+48.57\%)}\\
    \hline
    Proposed (YCbCr) & 61.23 \textcolor{red}{(+46.94\%)}& 62.31 \textcolor{red}{(+43.08\%)}& 234.44 \textcolor{red}{(+65.98\%)}& 242.25 \textcolor{red}{(+64.89\%)} & 118.64 \textcolor{red}{(+79.62\%)} & 226.41 \textcolor{red}{(+67.50\%)} & 229.95 \textcolor{red}{(+63.48\%)}& 167.89 \textcolor{red}{(+64.18\%)}\\
    \hline
    Proposed (HSL) & 85.43 \textcolor{red}{(+105.01\%)}& 87.99 \textcolor{red}{(+102.04\%)}& 256.71 \textcolor{red}{(+81.74\%)}& 277.38 \textcolor{red}{(+88.80\%)} & 131.46 \textcolor{red}{(+99.03\%)} & 237.94 \textcolor{red}{(+76.03\%)} & 249.89 \textcolor{red}{(+77.66\%)}& 189.54 \textcolor{red}{(+85.35\%)}\\
    \hline
    Proposed (HSV) & 64.36 \textcolor{red}{(+54.45\%)}& 65.37 \textcolor{red}{(+50.10\%)} & 237.17 \textcolor{red}{(+67.91\%)}& 241.10 \textcolor{red}{(+64.10\%)} & 120.21 \textcolor{red}{(+82.00\%)} & 224.03 \textcolor{red}{(+65.74\%)} & 232.85 \textcolor{red}{(+65.54\%)}& 169.30 \textcolor{red}{(+65.56\%)}\\
    \hline
    Proposed (CIE XYZ) & 51.12 \textcolor{red}{(+22.68\%)}& 53.45 \textcolor{red}{(+22.73\%)}& 215.11 \textcolor{red}{(+52.29\%)}& 220.92 \textcolor{red}{(+50.37\%)} & 108.71 \textcolor{red}{(+64.59\%)} & 215.65 \textcolor{red}{(+59.54\%)} & 212.06 \textcolor{red}{(+50.76\%)}& 153.86 \textcolor{red}{(+50.46\%)}\\
    \hline
    Proposed (CIE LAB) & 50.95 \textcolor{red}{(+22.27\%)}& 52.96 \textcolor{red}{(+21.61\%)}& 213.86 \textcolor{red}{(+51.41\%)}& 219.80 \textcolor{red}{(+49.61\%)} & 105.93 \textcolor{red}{(+60.38\%)} & 217.89 \textcolor{red}{(+61.20\%)} & 215.71 \textcolor{red}{(+53.36\%)}& 153.87 \textcolor{red}{(+50.47\%)}\\
    \hline
    \end{tabular}}
    \label{tab:perf_compare}
\end{table*}

\begin{figure}[htb!]
    \centerline{\includegraphics[width=\linewidth]{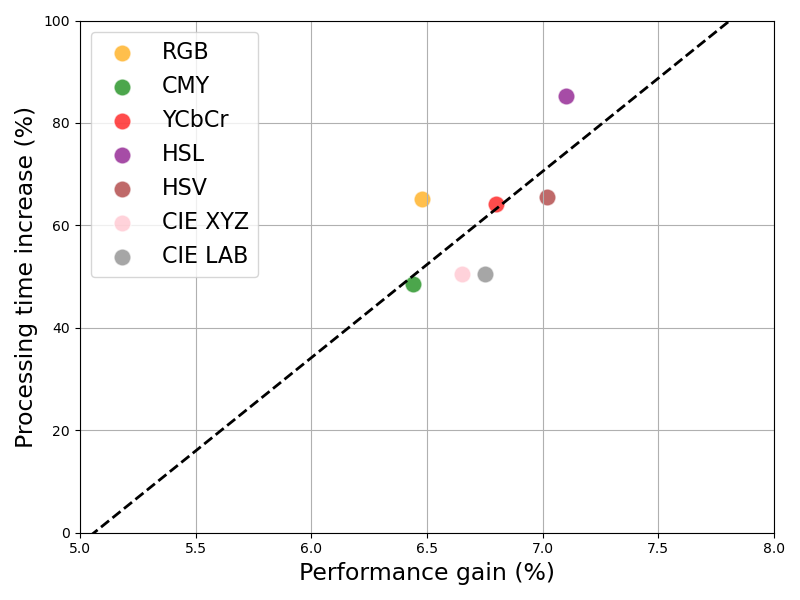}}
    \caption{Relationship between average gain in performance (higher is better) and average increase in processing time (lower is better) of different color models.}
    \label{fig:scatter_perf_avg}
\end{figure}

Our experimental results demonstrate that bit-plane coding can enhance the performance of SNNs. Despite the significant performance gains, the added computational complexity is a limitation of our method. As shown in Figure \ref{fig:scatter_perf_avg}, the average complexity increase in terms of processing time can be from 48.57\% to 85.35\%, which could be a drawback in some edge applications where resource utilization are of higher priority. Still, along with most spike coding methods utilized in surrogate training in SNN, both proposed method and baseline are still able to compute spike signal within linear time complexity \( \mathcal{O}(n) \)\cite{auge2021survey}. Empirical evaluation of processing speed and utilized memory could varied across datasets, attributed to factors such as input resolution, hardware specifications, and software library versions. Also, another limitation is that, similar to recent studies \cite{hu2021spiking, fang2021deep}, the experiments in our study are still based on software simulations. Implementations and experiments on hardware would be definitely needed to understand the actual complexity.
  
As our study is the first one to investigate bit plane coding in SNNs, it can be extended in different directions. First, a more extensive theoretical analysis is necessary to model the impact of bit plane coding on the performance of SNNs for different machine learning tasks or cross ANN-SNN benchmarking\cite{deng2020rethinking}. Second, it is obvious that the choice of color model may depend on the task in question, such as temporal-based segmentation\cite{patel2021spiking} , object recognition\cite{su2023deep} and knowledge distillation \cite{guo2023joint}. So, experiments in different application domains to understand the effects of the proposed method and color models for different tasks. Additionally for future works, the proposed method should be implemented by mapping proposed spike coding method on hardware platforms for deployment such as FPGA or Intel Loihi chip\cite{nunes2022spiking} and tested in practical scenarios such as low power edge devices\cite{zheng2021balancing}.   

\section{Conclusion}
In this paper, we have introduced a novel coding method for SNN that leverages bit planes of input image data. This approach proved to enhance the computational accuracy of SNN without increasing the model size. Through providing theoretically justified proof and extensive experimental validation, we have demonstrated the effectiveness of our coding strategy with multiple well known coding scheme across different grayscale and color datasets. This study is pioneering in the investigation of bit planes and color models within the context of SNN. We expect that the findings in this paper open new avenues for the development of more efficient and effective SNN models. In future work, we will further investigate the impacts of bit plane coding and color models in different application domains, such as medical analysis and object detection for traffic smart cameras. In addition, the proposed method will be implemented and tested on practical hardware platforms.

\section*{Acknowledgments}
We are sincerely thankful for the constructive criticisms and insights of reviewers for the betterment of our research.

\section*{Data availability}
Source code for our experiment is publicly available on GitHub (\url{https://github.com/luutn2002/bit-plane-snn}).

\bibliographystyle{IEEEtran}
\bibliography{refs}

@book{wani2020advances,
  title={Advances in deep learning},
  author={Wani, M Arif and Bhat, Farooq Ahmad and Afzal, Saduf and Khan, Asif Iqbal},
  year={2020},
  publisher={Springer}
}

@article{ngo2023survey,
  title={A survey of recent advances in quantum generative adversarial networks},
  author={Ngo, Tuan A and Nguyen, Tuyen and Thang, Truong Cong},
  journal={Electronics},
  volume={12},
  number={4},
  pages={856},
  year={2023},
  publisher={MDPI}
}

@article{masters2018revisiting,
  title={Revisiting small batch training for deep neural networks},
  author={Masters, Dominic and Luschi, Carlo},
  journal={arXiv preprint arXiv:1804.07612},
  year={2018}
}

@article{guo2023joint,
  title={Joint a-snn: Joint training of artificial and spiking neural networks via self-distillation and weight factorization},
  author={Guo, Yufei and Peng, Weihang and Chen, Yuanpei and Zhang, Liwen and Liu, Xiaode and Huang, Xuhui and Ma, Zhe},
  journal={Pattern Recognition},
  volume={142},
  pages={109639},
  year={2023},
  publisher={Elsevier}
}

@article{zheng2021balancing,
  title={Balancing the cost and performance trade-offs in SNN processors},
  author={Zheng, Huanliang and Guo, Yuhao and Yang, Xingyu and Xiao, Shanlin and Yu, Zhiyi},
  journal={IEEE Transactions on Circuits and Systems II: Express Briefs},
  volume={68},
  number={9},
  pages={3172--3176},
  year={2021},
  publisher={IEEE}
}

@article{patel2021spiking,
  title={A spiking neural network for image segmentation},
  author={Patel, Kinjal and Hunsberger, Eric and Batir, Sean and Eliasmith, Chris},
  journal={arXiv preprint arXiv:2106.08921},
  year={2021}
}

@inproceedings{su2023deep,
  title={Deep directly-trained spiking neural networks for object detection},
  author={Su, Qiaoyi and Chou, Yuhong and Hu, Yifan and Li, Jianing and Mei, Shijie and Zhang, Ziyang and Li, Guoqi},
  booktitle={Proceedings of the IEEE/CVF International Conference on Computer Vision},
  pages={6555--6565},
  year={2023}
}

@article{deng2020rethinking,
  title={Rethinking the performance comparison between SNNS and ANNS},
  author={Deng, Lei and Wu, Yujie and Hu, Xing and Liang, Ling and Ding, Yufei and Li, Guoqi and Zhao, Guangshe and Li, Peng and Xie, Yuan},
  journal={Neural networks},
  volume={121},
  pages={294--307},
  year={2020},
  publisher={Elsevier}
}

@article{luu2025accuracy,
  title={Accuracy-Robustness Trade Off via Spiking Neural Network Gradient Sparsity Trail},
  author={Luu, Nhan T},
  journal={arXiv preprint arXiv:2509.23762},
  year={2025}
}

@article{luu2025hybrid,
  title={Hybrid Layer-Wise ANN-SNN With Surrogate Spike Encoding-Decoding Structure},
  author={Luu, Nhan T and Luu, Duong T and Nam, Pham Ngoc and Thang, Truong Cong},
  journal={arXiv preprint arXiv:2509.24411},
  year={2025}
}

@article{zakon2004mathematical,
  title={Mathematical Analysis. Volume I/Zakon, Elias},
  author={Zakon, Elias},
  journal={West Lafayette, Indiana, USA},
  year={2004}
}

@inproceedings{luu2023blind,
  title={Blind image quality assessment with multimodal prompt learning},
  author={Luu, Nhan T and Onuoha, Chibuike and Thang, Truong Cong},
  booktitle={2023 IEEE 15th International Conference on Computational Intelligence and Communication Networks (CICN)},
  pages={614--618},
  year={2023},
  organization={IEEE}
}

@inproceedings{radford2021learning,
  title={Learning transferable visual models from natural language supervision},
  author={Radford, Alec and Kim, Jong Wook and Hallacy, Chris and Ramesh, Aditya and Goh, Gabriel and Agarwal, Sandhini and Sastry, Girish and Askell, Amanda and Mishkin, Pamela and Clark, Jack and others},
  booktitle={International conference on machine learning},
  pages={8748--8763},
  year={2021},
  organization={PmLR}
}

@article{nguyen2022evaluation,
  title={An evaluation of hardware-efficient quantum neural networks for image data classification},
  author={Nguyen, Tuyen and Paik, Incheon and Watanobe, Yutaka and Thang, Truong Cong},
  journal={Electronics},
  volume={11},
  number={3},
  pages={437},
  year={2022},
  publisher={MDPI}
}

@inproceedings{nguyen2022quantum,
  title={Quantum machine learning with quantum image representations},
  author={Nguyen, Tuyen and Paik, Incheon and Sagawa, Hiroyuki and Thang, Truong Cong},
  booktitle={2022 IEEE International Conference on Quantum Computing and Engineering (QCE)},
  pages={851--854},
  year={2022},
  organization={IEEE}
}

@inproceedings{luu2024,
  title={Improvement of Spiking Neural Network With Bit Plane Coding},
  author={Luu, Nhan and Luu, Duong and Nam, Pham Ngoc and Thang, Truong Cong},
  booktitle={2024 IEEE 16th International Conference on Computational Intelligence and Communication Networks (CICN)},
  year={2024},
  organization={IEEE}
}

@article{nguyen2021review,
  title={A review of algorithms and hardware implementations for spiking neural networks},
  author={Nguyen, Duy-Anh and Tran, Xuan-Tu and Iacopi, Francesca},
  journal={Journal of Low Power Electronics and Applications},
  volume={11},
  number={2},
  pages={23},
  year={2021},
  publisher={MDPI}
}

@inproceedings{li2021free,
  title={A free lunch from ANN: Towards efficient, accurate spiking neural networks calibration},
  author={Li, Yuhang and Deng, Shikuang and Dong, Xin and Gong, Ruihao and Gu, Shi},
  booktitle={International conference on machine learning},
  pages={6316--6325},
  year={2021},
  organization={PMLR}
}

@article{deng2021optimal,
  title={Optimal conversion of conventional artificial neural networks to spiking neural networks},
  author={Deng, Shikuang and Gu, Shi},
  journal={arXiv preprint arXiv:2103.00476},
  year={2021}
}

@article{auge2021survey,
  title={A survey of encoding techniques for signal processing in spiking neural networks},
  author={Auge, Daniel and Hille, Julian and Mueller, Etienne and Knoll, Alois},
  journal={Neural Processing Letters},
  volume={53},
  number={6},
  pages={4693--4710},
  year={2021},
  publisher={Springer}
}

@inproceedings{blouw2019benchmarking,
  title={Benchmarking keyword spotting efficiency on neuromorphic hardware},
  author={Blouw, Peter and Choo, Xuan and Hunsberger, Eric and Eliasmith, Chris},
  booktitle={Proceedings of the 7th annual neuro-inspired computational elements workshop},
  pages={1--8},
  year={2019}
}

@article{nunes2022spiking,
  title={Spiking neural networks: A survey},
  author={Nunes, Joao D and Carvalho, Marcelo and Carneiro, Diogo and Cardoso, Jaime S},
  journal={IEEE Access},
  volume={10},
  pages={60738--60764},
  year={2022},
  publisher={IEEE}
}

@article{neftci2019surrogate,
  title={Surrogate gradient learning in spiking neural networks: Bringing the power of gradient-based optimization to spiking neural networks},
  author={Neftci, Emre O and Mostafa, Hesham and Zenke, Friedemann},
  journal={IEEE Signal Processing Magazine},
  volume={36},
  number={6},
  pages={51--63},
  year={2019},
  publisher={IEEE}
}

@inproceedings{srinivasan2017spike,
  title={Spike timing dependent plasticity based enhanced self-learning for efficient pattern recognition in spiking neural networks},
  author={Srinivasan, Gopalakrishnan and Roy, Sourjya and Raghunathan, Vijay and Roy, Kaushik},
  booktitle={2017 International Joint Conference on Neural Networks (IJCNN)},
  pages={1847--1854},
  year={2017},
  organization={IEEE}
}

@article{liu2021sstdp,
  title={SSTDP: Supervised spike timing dependent plasticity for efficient spiking neural network training},
  author={Liu, Fangxin and Zhao, Wenbo and Chen, Yongbiao and Wang, Zongwu and Yang, Tao and Jiang, Li},
  journal={Frontiers in Neuroscience},
  volume={15},
  pages={756876},
  year={2021},
  publisher={Frontiers Media SA}
}

@inproceedings{gowda2019colornet,
  title={ColorNet: Investigating the importance of color spaces for image classification},
  author={Gowda, Shreyank N and Yuan, Chun},
  booktitle={Computer Vision--ACCV 2018: 14th Asian Conference on Computer Vision, Perth, Australia, December 2--6, 2018, Revised Selected Papers, Part IV 14},
  pages={581--596},
  year={2019},
  organization={Springer}
}

@inproceedings{taipalmaa2020different,
  title={Different color spaces in deep learning-based water segmentation for autonomous marine operations},
  author={Taipalmaa, Jussi and Passalis, Nikolaos and Raitoharju, Jenni},
  booktitle={2020 IEEE international conference on image processing (ICIP)},
  pages={3169--3173},
  year={2020},
  organization={IEEE}
}

@inproceedings{vorabbi2023input,
  title={Input Layer Binarization with Bit-Plane Encoding},
  author={Vorabbi, Lorenzo and Maltoni, Davide and Santi, Stefano},
  booktitle={International Conference on Artificial Neural Networks},
  pages={395--406},
  year={2023},
  organization={Springer}
}

@article{rajendran2019low,
  title={Low-power neuromorphic hardware for signal processing applications: A review of architectural and system-level design approaches},
  author={Rajendran, Bipin and Sebastian, Abu and Schmuker, Michael and Srinivasa, Narayan and Eleftheriou, Evangelos},
  journal={IEEE Signal Processing Magazine},
  volume={36},
  number={6},
  pages={97--110},
  year={2019},
  publisher={IEEE}
}

@article{maass1997networks,
  title={Networks of spiking neurons: the third generation of neural network models},
  author={Maass, Wolfgang},
  journal={Neural networks},
  volume={10},
  number={9},
  pages={1659--1671},
  year={1997},
  publisher={Elsevier}
}

@book{gerstner2002spiking,
  title={Spiking neuron models: Single neurons, populations, plasticity},
  author={Gerstner, Wulfram and Kistler, Werner M},
  year={2002},
  publisher={Cambridge university press}
}

@article{kim2018deep,
  title={Deep neural networks with weighted spikes},
  author={Kim, Jaehyun and Kim, Heesu and Huh, Subin and Lee, Jinho and Choi, Kiyoung},
  journal={Neurocomputing},
  volume={311},
  pages={373--386},
  year={2018},
  publisher={Elsevier}
}

@article{heeger2000poisson,
  title={Poisson model of spike generation},
  author={Heeger, David and others},
  journal={Handout, University of Standford},
  volume={5},
  number={1-13},
  pages={76},
  year={2000}
}

@article{ibraheem2012understanding,
  title={Understanding color models: a review},
  author={Ibraheem, Noor A and Hasan, Mokhtar M and Khan, Rafiqul Z and Mishra, Pramod K},
  journal={ARPN Journal of science and technology},
  volume={2},
  number={3},
  pages={265--275},
  year={2012},
  publisher={Citeseer}
}

@article{paszke2019pytorch,
  title={Pytorch: An imperative style, high-performance deep learning library},
  author={Paszke, Adam and Gross, Sam and Massa, Francisco and Lerer, Adam and Bradbury, James and Chanan, Gregory and Killeen, Trevor and Lin, Zeming and Gimelshein, Natalia and Antiga, Luca and others},
  journal={Advances in neural information processing systems},
  volume={32},
  year={2019}
}

@article{plataniotis2001color,
  title={Color image processing and applications},
  author={Plataniotis, Konstantinos N},
  journal={Measurement Science and Technology},
  volume={12},
  number={2},
  pages={222--222},
  year={2001}
}

@article{kingma2014adam,
  title={Adam: A method for stochastic optimization},
  author={Kingma, Diederik P and Ba, Jimmy},
  journal={arXiv preprint arXiv:1412.6980},
  year={2014}
}

@inproceedings{eriba2019kornia,
  author    = {E. Riba and D. Mishkin and D. Ponsa and E. Rublee and G. Bradski},
  title     = {Kornia: an Open Source Differentiable Computer Vision Library for PyTorch},
  booktitle = {Winter Conference on Applications of Computer Vision},
  year      = {2020},
  url       = {https://arxiv.org/pdf/1910.02190.pdf}
}

@article{
doi:10.1126/sciadv.adi1480,
author = {Wei Fang  and Yanqi Chen  and Jianhao Ding  and Zhaofei Yu  and Timothée Masquelier  and Ding Chen  and Liwei Huang  and Huihui Zhou  and Guoqi Li  and Yonghong Tian },
title = {SpikingJelly: An open-source machine learning infrastructure platform for spike-based intelligence},
journal = {Science Advances},
volume = {9},
number = {40},
pages = {eadi1480},
year = {2023},
doi = {10.1126/sciadv.adi1480},
URL = {https://www.science.org/doi/abs/10.1126/sciadv.adi1480},
eprint = {https://www.science.org/doi/pdf/10.1126/sciadv.adi1480}}

@article{fang2021deep,
  title={Deep residual learning in spiking neural networks},
  author={Fang, Wei and Yu, Zhaofei and Chen, Yanqi and Huang, Tiejun and Masquelier, Timoth{\'e}e and Tian, Yonghong},
  journal={Advances in Neural Information Processing Systems},
  volume={34},
  pages={21056--21069},
  year={2021}
}

@article{alex2009learning,
  title={Learning multiple layers of features from tiny images},
  author={Alex, Krizhevsky},
  journal={https://www. cs. toronto. edu/kriz/learning-features-2009-TR. pdf},
  year={2009}
}

@article{helber2019eurosat,
  title={Eurosat: A novel dataset and deep learning benchmark for land use and land cover classification},
  author={Helber, Patrick and Bischke, Benjamin and Dengel, Andreas and Borth, Damian},
  journal={IEEE Journal of Selected Topics in Applied Earth Observations and Remote Sensing},
  volume={12},
  number={7},
  pages={2217--2226},
  year={2019},
  publisher={IEEE}
}

@article{dampfhoffer2022snns,
  title={Are SNNs really more energy-efficient than ANNs? An in-depth hardware-aware study},
  author={Dampfhoffer, Manon and Mesquida, Thomas and Valentian, Alexandre and Anghel, Lorena},
  journal={IEEE Transactions on Emerging Topics in Computational Intelligence},
  volume={7},
  number={3},
  pages={731--741},
  year={2022},
  publisher={IEEE}
}

@article{rumelhart1986learning,
  title={Learning representations by back-propagating errors},
  author={Rumelhart, David E and Hinton, Geoffrey E and Williams, Ronald J},
  journal={nature},
  volume={323},
  number={6088},
  pages={533--536},
  year={1986},
  publisher={Nature Publishing Group UK London}
}

@article{rueckauer2017conversion,
  title={Conversion of continuous-valued deep networks to efficient event-driven networks for image classification},
  author={Rueckauer, Bodo and Lungu, Iulia-Alexandra and Hu, Yuhuang and Pfeiffer, Michael and Liu, Shih-Chii},
  journal={Frontiers in neuroscience},
  volume={11},
  pages={682},
  year={2017},
  publisher={Frontiers Media SA}
}

@article{lee2018training,
  title={Training deep spiking convolutional neural networks with STDP-based unsupervised pre-training followed by supervised fine-tuning},
  author={Lee, Chankyu and Panda, Priyadarshini and Srinivasan, Gopalakrishnan and Roy, Kaushik},
  journal={Frontiers in neuroscience},
  volume={12},
  pages={435},
  year={2018},
  publisher={Frontiers Media SA}
}

@article{merolla2014million,
  title={A million spiking-neuron integrated circuit with a scalable communication network and interface},
  author={Merolla, Paul A and Arthur, John V and Alvarez-Icaza, Rodrigo and Cassidy, Andrew S and Sawada, Jun and Akopyan, Filipp and Jackson, Bryan L and Imam, Nabil and Guo, Chen and Nakamura, Yutaka and others},
  journal={Science},
  volume={345},
  number={6197},
  pages={668--673},
  year={2014},
  publisher={American Association for the Advancement of Science}
}

@misc{li_andreeto_ranzato_perona_2022, title={Caltech 101}, DOI={10.22002/D1.20086}, abstractNote={Pictures of objects belonging to 101 categories. About 40 to 800 images per category. Most categories have about 50 images. Collected in September 2003 by Fei-Fei Li, Marco Andreetto, and Marc'Aurelio Ranzato. The size of each image is roughly 300 x 200 pixels. We have carefully clicked outlines of each object in these pictures, these are included under the 'Annotations.tar'. There is also a MATLAB script to view the annotations, 'show_annotations.m'.}, publisher={CaltechDATA}, author={Li, Fei-Fei and Andreeto, Marco and Ranzato, Marc'Aurelio and Perona, Pietro}, year={2022}, month={Apr} }

@misc{griffin_holub_perona_2022, title={Caltech 256}, DOI={10.22002/D1.20087}, abstractNote={We introduce a challenging set of 256 object categories containing a total of 30607 images. The original Caltech-101 was collected by choosing a set of object categories, downloading examples from Google Images and then manually screening out all images that did not fit the category. Caltech-256 is collected in a similar manner with several improvements: a) the number of categories is more than doubled, b) the minimum number of images in any category is increased from 31 to 80, c) artifacts due to image rotation are avoided and d) a new and larger clutter category is introduced for testing background rejection. We suggest several testing paradigms to measure classification performance, then benchmark the dataset using two simple metrics as well as a state-of-the-art spatial pyramid matching algorithm. Finally we use the clutter category to train an interest detector which rejects uninformative background regions.}, publisher={CaltechDATA}, author={Griffin, Gregory and Holub, Alex and Perona, Pietro}, year={2022}, month={Apr} }

@misc{imagenette,
    author    = "Jeremy Howard",
    title     = "Imagenette",
    url       = "https://github.com/fastai/imagenette/"
}

@inproceedings{deng2009imagenet,
  title={Imagenet: A large-scale hierarchical image database},
  author={Deng, Jia and Dong, Wei and Socher, Richard and Li, Li-Jia and Li, Kai and Fei-Fei, Li},
  booktitle={2009 IEEE conference on computer vision and pattern recognition},
  pages={248--255},
  year={2009},
  organization={Ieee}
}

@inproceedings{food101,
  title = {Food-101 -- Mining Discriminative Components with Random Forests},
  author = {Bossard, Lukas and Guillaumin, Matthieu and Van Gool, Luc},
  booktitle = {European Conference on Computer Vision},
  year = {2014}
}

@article{deng2012mnist,
  title={The mnist database of handwritten digit images for machine learning research [best of the web]},
  author={Deng, Li},
  journal={IEEE signal processing magazine},
  volume={29},
  number={6},
  pages={141--142},
  year={2012},
  publisher={IEEE}
}

@article{xiao2017fashion,
  title={Fashion-mnist: a novel image dataset for benchmarking machine learning algorithms},
  author={Xiao, Han and Rasul, Kashif and Vollgraf, Roland},
  journal={arXiv preprint arXiv:1708.07747},
  year={2017}
}

@online{clanuwat2018deep,
  author       = {Tarin Clanuwat and Mikel Bober-Irizar and Asanobu Kitamoto and Alex Lamb and Kazuaki Yamamoto and David Ha},
  title        = {Deep Learning for Classical Japanese Literature},
  date         = {2018-12-03},
  year         = {2018},
  eprintclass  = {cs.CV},
  eprinttype   = {arXiv},
  eprint       = {cs.CV/1812.01718},
}

@article{abbott1999lapicque,
  title={Lapicque’s introduction of the integrate-and-fire model neuron (1907)},
  author={Abbott, Larry F},
  journal={Brain research bulletin},
  volume={50},
  number={5-6},
  pages={303--304},
  year={1999},
  publisher={Citeseer}
}

@inproceedings{diehl2015fast,
  title={Fast-classifying, high-accuracy spiking deep networks through weight and threshold balancing},
  author={Diehl, Peter U and Neil, Daniel and Binas, Jonathan and Cook, Matthew and Liu, Shih-Chii and Pfeiffer, Michael},
  booktitle={2015 International joint conference on neural networks (IJCNN)},
  pages={1--8},
  year={2015},
  organization={ieee}
}

@article{hu2021spiking,
  title={Spiking deep residual networks},
  author={Hu, Yangfan and Tang, Huajin and Pan, Gang},
  journal={IEEE Transactions on Neural Networks and Learning Systems},
  volume={34},
  number={8},
  pages={5200--5205},
  year={2021},
  publisher={IEEE}
}

@inproceedings{he2016deep,
  title={Deep residual learning for image recognition},
  author={He, Kaiming and Zhang, Xiangyu and Ren, Shaoqing and Sun, Jian},
  booktitle={Proceedings of the IEEE conference on computer vision and pattern recognition},
  pages={770--778},
  year={2016}
}

@article{kim2018efficient,
  title={An Efficient Color Space for Deep-Learning Based Traffic Light Recognition},
  author={Kim, Hyun-Koo and Park, Ju H and Jung, Ho-Youl},
  journal={Journal of Advanced Transportation},
  volume={2018},
  number={1},
  pages={2365414},
  year={2018},
  publisher={Wiley Online Library}
}

@article{chyad2025exploring,
  title={Exploring adversarial deep learning for fusion in multi-color channel skin detection applications},
  author={Chyad, Mohammed and Zaidan, BB and Zaidan, AA and Pilehkouhi, Hossein and Aalaa, Roqia and Qahtan, Sarah and Alsattar, Hassan A and Pamucar, Dragan and Simic, Vladimir},
  journal={Information Fusion},
  volume={114},
  pages={102632},
  year={2025},
  publisher={Elsevier}
}

@inproceedings{zhang2019tdsnn,
  title={Tdsnn: From deep neural networks to deep spike neural networks with temporal-coding},
  author={Zhang, Lei and Zhou, Shengyuan and Zhi, Tian and Du, Zidong and Chen, Yunji},
  booktitle={Proceedings of the AAAI conference on artificial intelligence},
  volume={33},
  number={01},
  pages={1319--1326},
  year={2019}
}

@article{liu2022defending,
  title={Defending against Adversarial Attacks in Deep Learning with Robust Auxiliary Classifiers Utilizing Bit-plane Slicing},
  author={Liu, Yuan and Dong, Jinxin and Zhou, Pingqiang},
  journal={ACM Journal on Emerging Technologies in Computing Systems (JETC)},
  volume={18},
  number={3},
  pages={1--17},
  year={2022},
  publisher={ACM New York, NY}
}

@inproceedings{chen2019breast,
  title={Breast cancer image classification based on CNN and bit-plane slicing},
  author={Chen, Guoming and Chen, Yongchang and Yuan, Zeduo and Lu, Xuming and Zhu, Xiongyong and Li, Wanyi},
  booktitle={2019 International Conference on Medical Imaging Physics and Engineering (ICMIPE)},
  pages={1--4},
  year={2019},
  organization={IEEE}
}

@article{eshraghian2023training,
  title={Training spiking neural networks using lessons from deep learning},
  author={Eshraghian, Jason K and Ward, Max and Neftci, Emre O and Wang, Xinxin and Lenz, Gregor and Dwivedi, Girish and Bennamoun, Mohammed and Jeong, Doo Seok and Lu, Wei D},
  journal={Proceedings of the IEEE},
  year={2023},
  publisher={IEEE}
}

@book{abelson1996structure,
  title={Structure and interpretation of computer programs},
  author={Abelson, Harold and Sussman, Gerald Jay},
  year={1996},
  publisher={The MIT Press}
}

@article{huh2018gradient,
  title={Gradient descent for spiking neural networks},
  author={Huh, Dongsung and Sejnowski, Terrence J},
  journal={Advances in neural information processing systems},
  volume={31},
  year={2018}
}

@inproceedings{shalev2017failures,
  title={Failures of gradient-based deep learning},
  author={Shalev-Shwartz, Shai and Shamir, Ohad and Shammah, Shaked},
  booktitle={International Conference on Machine Learning},
  pages={3067--3075},
  year={2017},
  organization={PMLR}
}

@article{ghadimi2013stochastic,
  title={Stochastic first-and zeroth-order methods for nonconvex stochastic programming},
  author={Ghadimi, Saeed and Lan, Guanghui},
  journal={SIAM journal on optimization},
  volume={23},
  number={4},
  pages={2341--2368},
  year={2013},
  publisher={SIAM}
}

@inproceedings{cochran1934distribution,
  title={The distribution of quadratic forms in a normal system, with applications to the analysis of covariance},
  author={Cochran, William G},
  booktitle={Mathematical Proceedings of the Cambridge Philosophical Society},
  volume={30},
  number={2},
  pages={178--191},
  year={1934},
  organization={Cambridge University Press}
}

@article{laurent2000adaptive,
  title={Adaptive estimation of a quadratic functional by model selection},
  author={Laurent, Beatrice and Massart, Pascal},
  journal={Annals of statistics},
  pages={1302--1338},
  year={2000},
  publisher={JSTOR}
}

@inproceedings{littlewood1939number,
  title={On the number of real roots of a random algebraic equation. II},
  author={Littlewood, John Edensor and Offord, Albert C},
  booktitle={Mathematical proceedings of the Cambridge philosophical society},
  volume={35},
  number={2},
  pages={133--148},
  year={1939},
  organization={Cambridge University Press}
}

\begin{appendices} \label{apdx}
\section{Basis for stochastic theorem reusability in SNN surrogate training} \label{reusability}

As discussed in Section \ref{surrogate_grad_bg}, surrogate gradient methods attempt to adapt deep learning's gradient optimization techniques to SNNs, maintaining a close connection to traditional deep learning and convex optimization. Given well-established conditions and appropriate modifications, the theoretical principles from these fields can be leveraged to explain the behavior of surrogate gradient methods. To prove this viewpoint we consider a classical stochastic optimization problem associated with learning a target function $h$ can be formulated as:
\begin{equation}
    \min_{\mathbf{w}} F_h(\mathbf{w}) = \mathbb{E}_{\mathbf{x}} \left[ \ell(p_{\mathbf{w}}(\mathbf{x}), h(\mathbf{x})) \right]
\end{equation}
where $\ell$ is a loss function, $\mathbf{x}$ represents stochastic inputs (assumed to be vectors in Euclidean space), and $p_{\mathbf{w}}$ is a differentiable predictor parameterized by $\mathbf{w}$ (e.g., a traditional ANN of a specific architecture). It is assumed that $E_x[\left\|\frac{\partial}{\partial{\mathbf{w}}}p_{{\mathbf{w}}}\right\|^2] \leq G(\mathbf{w})^2$ for some scalar $G(\mathbf{w})$, and that $F$ is differentiable. The variance of the given cost function can be quantified as:
\begin{equation}
    Var(\mathcal{H},F,\mathbf{w})=\mathbb{E}_{h}\left\|\nabla F_h(\mathbf{w})-\mathbb{E}_{h'}\nabla F_{h'}(\mathbf{w})\right\|^2
\end{equation}

Now, we consider an example theorem along with it's auxiliary lemmas derived from a theoretical analysis of a gradient-based optimization experiment\cite{shalev2017failures}:
\begin{theorem}[Shalev et al., 2017]
\label{var_theorem}
Let \(\mathbf{x}_{1}^{k}\) denote a \(k\)-tuple \((\mathbf{x}_{1},\ldots,\mathbf{x}_{k})\) of input instances, and assume that each \(\mathbf{x}_{l}\) is i.i.d. standard Gaussian in \(\mathbb{R}^{d}\). Define
\[
h_{\mathbf{u}}(\mathbf{x}_{1}^{k})=\prod_{l=1}^{k}\text{sign}(\mathbf{u}^{\top}\mathbf{x}_{l}),
\]
and the objective (w.r.t. some predictor \(p_{\mathbf{w}}\) parameterized by \(\mathbf{w}\))
\[
F(\mathbf{w})=\mathop{\mathbb{E}}\limits_{\mathbf{x}_{1}^{k}}\left[\ell(p_{\mathbf{w}}(\mathbf{x}_{1}^{k}),h_{\mathbf{u}}(\mathbf{x}_{1}^{k}))\right].
\]
Where the loss function \(\ell\) is either the square loss \(\ell(\hat{y},y)=\frac{1}{2}(\hat{y}-y)^{2}\) or a classification loss of the form \(\ell(\hat{y},y)=r(\hat{y}\cdot y)\) for some \(1\)-Lipschitz function \(r\). 

Fix some \(\mathbf{w}\), and suppose that \(p_{\mathbf{w}}(\mathbf{x})\) is differentiable w.r.t. \(\mathbf{w}\) and satisfies \(\mathop{\mathbb{E}}\limits_{\mathbf{x}_{1}^{k}}\left[\|\frac{\partial}{\partial \mathbf{w}}p_{\mathbf{w}}(\mathbf{x}_{1}^{k})\|^{2}\right]\leq G(\mathbf{w})^{2}\). Then if \(\mathcal{H}=\{h_{\mathbf{u}}:\mathbf{u}\in\mathbb{R}^{d},\|\mathbf{u}\|=1\}\), then
\[
\textrm{Var}(\mathcal{H},F,\mathbf{w})\;\leq\;G(\mathbf{w})^{2}\cdot O\left( \sqrt{\frac{k\log(d)}{d}}\right)^{k}.
\]
\end{theorem}
\begin{lemma}[Shalev et al., 2017]
\label{lemma_1}
Let \(h_{1},\ldots,h_{n}\) be real-valued functions on some Euclidean space, which belong to some weighted \(L_{2}\) space. Suppose that \(\|h_{i}\|_{L_{2}}=1\) and \(\max_{i\neq j}|\langle h_{i},h_{j}\rangle_{L_{2}}|\leq c\). Then for any function \(g\) on the same domain,
\[
\frac{1}{n}\sum_{i=1}^{n}\langle h_{i},g\rangle_{L_{2}}^{2}\,\leq\,\|g\|_{L_{2}}^{2}\left(\frac{1}{n}+c\right).
\]
\end{lemma}
\begin{lemma}[Shalev et al., 2017]
\label{lemma_2}
If \(\mathbf{w},\mathbf{v}\) are two unit vectors in \(\mathbb{R}^{d}\), and \(\mathbf{x}\) is a standard Gaussian random vector, then
\[
\left|\mathbb{E}_{\mathbf{x}}\left[\text{sign}(\mathbf{w}^{\top}\mathbf{x})\,\text{sign}(\mathbf{v}^{\top}\mathbf{x})\right]\right|\leq|\langle\mathbf{w},\mathbf{v}\rangle|
\]
\end{lemma}

Theorem \ref{var_theorem} holds true as long as Lemma \ref{lemma_1} and Lemma \ref{lemma_2} remain valid. There are two primary approaches to generalizing this theorem for surrogate gradient methods:

\begin{itemize}
    \item Demonstrating that as the distribution of the input $\mathbf{x}$ have a distribution similar to a Bernoulli distribution, the lemmas remain valid, thereby ensuring the theorem holds.
    \item Reformulating the surrogate gradient problem as a conventional stochastic optimization problem.
\end{itemize}

We first consider the first approach. Suppose each entry $\mathbf{x_i}$ of the vector $\mathbf{x}$ follows a Bernoulli distribution with parameter $p_i$, so that the joint probability of $\mathbf{x}$ is given by:
\[
Pr(\mathbf{x}) = \prod_{i} p_{i}^{x_{i}} (1 - p_{i})^{1 - x_{i}}.
\]
Since the components of Lemma \ref{lemma_1} do not depend on the distribution of $\mathbf{x}$, the lemma remains unchanged. However, Lemma \ref{lemma_2} no longer necessarily holds, as any multiplication involving $\mathbf{x}$ can be interpreted as a $d$-dimensional Littlewood–Offord problem\cite{littlewood1939number}, which only admits a probabilistic upper bound rather than a deterministic one. Consequently, Lemma \ref{lemma_2} becomes unstable and holds only under the additional assumption that the unit vectors also follow a standard Gaussian distribution, i.e., $\mathbf{w}, \mathbf{v} \sim \mathcal{N}(0,1)$. 

Under this assumption, \( w^T x \) and \( v^T x \) are jointly zero-mean Gaussian with variance 1 and covariance \( \mathbb{E}[w^T x x^T v] = w^T v \). Thus,  
\begin{equation}
    \begin{aligned}
        \mathbb{E} \left[ \text{sign}(w^T x) \text{sign}(v^T x) \right] &= 2 \Pr(w^T x \geq 0, v^T x \geq 0) \\
        &- 2 \Pr(w^T x \geq 0, v^T x \leq 0).\\
    \end{aligned}
\end{equation}
Using the quadrant probability of bivariate normal distributions, this simplifies to:
\begin{equation}
    \begin{aligned}
        &2 \left( \frac{1}{4} + \frac{\sin^{-1}(w^T v)}{2\pi} \right) - 2 \left( \frac{\cos^{-1}(w^T v)}{2\pi} \right) \\
        &= \frac{2 \sin^{-1}(w^T v)}{\pi}.\\
    \end{aligned}
\end{equation}
The absolute value of this expression is upper bounded by \( |w^T v| \), thereby establishing the stability of Lemma \ref{lemma_2} as long as the condition of $\mathbf{w}, \mathbf{v} \sim \mathcal{N}(0,1)$ held truth. However, proving each theorem along with its auxiliary lemmas under different input distributions shift would be inefficient since the theorem only hold within strictly constructed condition.

Alternatively, we can approach this by reformulating surrogate gradient method as a traditional stochastic optimization problem. In this case, our objective is defined as:
\begin{equation}
    \begin{aligned}
        \min_{\mathbf{w}} F_h(\mathbf{w}) &= \mathbb{E}_{\mathbf{x}} \left[ \ell(P_{\mathbf{w}}(\mathbf{x}), h(\mathbf{x})) \right], \\
        P_{\mathbf{w}}(\mathbf{x}) &= \mathbb{E}_{T} \left[ p_{\mathbf{w}}(e(\mathbf{x})) \right],
    \end{aligned}
\end{equation}
where the surrogate gradient model $P_{\mathbf{w}}(\mathbf{x})$ encodes the input $\mathbf{x}$ into a spike signal using an encoding function $e(\mathbf{x})$ and computes the expected output signal across the time dimension $T$ to generate the final stochastic prediction. Using this reformulation, we ensure that as long as the encoding function $e(\mathbf{x})$ is differentiable, all conventional deep learning theorems remain applicable to the surrogate gradient model. If $e(\mathbf{x})$ is not differentiable, the theorem holds only if its components and auxiliary lemmas do not require gradient computation with respect to $\mathbf{x}$. In similar manner with Equation \ref{eq:snr}, gradient of coding function $\frac{\partial e(\mathbf{x})}{\partial\mathbf{x}}$ is not required and Equation \ref{eq:snr} can be apply to surrogate SNN model as is. 

\section{Proof for SNR inequality} \label{ineq_proof}

Firstly, we will provide some background knowledge to assist our mathematical derivation of Equation \ref{eq:snr_ineq}. Given a SNN model that employ hard-reset IF neurons with backpropagate through time (BPTT)\cite{fang2021deep, eshraghian2023training, doi:10.1126/sciadv.adi1480, luu2025hybrid, luu2025accuracy}, optimization scheme is defined as:
\begin{equation}
    \begin{aligned}
        & \begin{aligned}
            u_i^{(l)}(t) & = \bigl(1 - s_i^{(l)}(t-1)\bigr)\,u_i^{(l)}(t-1)\\
            & + W^{(l)} s^{(l-1)}(t) + b_i^{(l)},\\
        \end{aligned}\\
        &s_i^{(l)}(t) = \Theta\left(u_i^{(l)}(t), V_{\text{th}}\right)\\
        & \Theta(x, V_{th}) =
        \begin{cases}
        1 & \text{if } x - V_{th} \geq 0 \\
        0 & \text{otherwise}
        \end{cases} \\
        & \frac{\partial \mathcal{L}}{\partial W^{(l)}} = \sum_{t=1}^{T} \frac{\partial \mathcal{L}}{\partial s^{(l)}(t)} \cdot \frac{\partial s^{(l)}(t)}{\partial u^{(l)}(t)} \cdot \frac{\partial u^{(l)}(t)}{\partial W^{(l)}}\\ 
    \end{aligned}
    \label{eq:bptt}
\end{equation}
where:
\begin{itemize}
    \item $u_i^{(l)}(t)$ is the membrane potential of neuron $i$ in layer $l$ at time $t \in T$,
    \item $s_i^{(l)}(t)$ is the spike output of neuron $i$, similarly with prior layer signal $s_i^{(l-1)}(t)$,
    \item $V_{\text{th}}$ is the threshold voltage,
    \item synaptic weight $W^{(l)}$ and bias $b_i^{(l)}$,
    \item $\Theta$ is the Heaviside function,
    \item loss function $\mathcal{L}$ and it's gradient with respect to weight $\frac{\partial \mathcal{L}}{\partial W^{(l)}}$.
\end{itemize}

As noted from Appendix \ref{reusability}, SNR expression denoted in Equation \ref{eq:snr} \cite{shalev2017failures} is adaptable to surrogate SNN model by substitute results from Equation \ref{eq:bptt} as:
\begin{equation}
    \begin{aligned}
        & SNR(\delta(\mathbf{x})) = \frac{\left\| \mathbb{E}_{\mathbf{x}} \left[ \delta(\mathbf{x}) \right] \right\|^2_2}{\mathbb{E}_{\mathbf{x}}\left\|\delta(\mathbf{x})-\mathbb{E}_{\mathbf{x}}\left[\delta(\mathbf{x})\right]\right\|^2_2},\\
        & \begin{aligned}
            \delta(\mathbf{x}) & = h(\mathbf{x}) \cdot g^{T}_\mathbf{w}(\mathbf{x})\\
            & = h(\mathbf{x}) \cdot \sum_{t=1}^{T} \frac{\partial s^{(l)}(t)}{\partial u^{(l)}(t)} \cdot \frac{\partial u^{(l)}(t)}{\partial W^{(l)}}\\ \\
        \end{aligned}
    \end{aligned}
\end{equation}
where $g_{{\mathbf{w}}}({\mathbf{x}}) =\frac{\partial}{\partial{\mathbf{w}}}p_{{\mathbf{w}}}({\mathbf{x}})$ is the Jacobian of model $p_{{\mathbf{w}}}({\mathbf{x}})$ with respect to model weight $\mathbf{w}$. Under the case of signal concatenation with bit plane coding of length $N_{bit}$, we would have $\delta^{bit}(\mathbf{x})$:
\begin{equation}
    \begin{aligned} 
        \delta^{bit}(\mathbf{x}) & = h(\mathbf{x}) \cdot \Big( \sum_{t=1}^{T} \frac{\partial s^{(l)}(t)}{\partial u^{(l)}(t)} \cdot \frac{\partial u^{(l)}(t)}{\partial W^{(l)}}\\
        & + \sum_{t=1}^{N_{bit}} \frac{\partial s^{(l)}(t)}{\partial u^{(l)}(t)} \cdot \frac{\partial u^{(l)}(t)}{\partial W^{(l)}}\Big)\\
        & = h(\mathbf{x}) \cdot g^{T}_\mathbf{w}(\mathbf{x}) + h(\mathbf{x}) \cdot g^{bit}_\mathbf{w}(\mathbf{x})\\
        & = h(\mathbf{x}) \cdot g^{T+bit}_\mathbf{w}(\mathbf{x})
    \end{aligned}
\end{equation}
Equation \ref{eq:snr_ineq} hold true if the following theorem is true:
\begin{theorem}
\label{main_theorem}
Assume that bit channels concatenated gradient $g^{bit}_\mathbf{w}(\mathbf{x})$ does not explode nor vanish, bounded as:
\[
G(\mathbf{w})^2 > \left\|  \mathbb{E}_{\mathbf{x}}\left[g^{bit}_\mathbf{w}(\mathbf{x})\right]\right\|_\infty > 0
\] 
for some scalar $G(\mathbf{w})$, where both gradient of baseline SNN $g^T_\mathbf{w}(\mathbf{x})$ and bit plane code concatenated SNN $g^{bit}_\mathbf{w}(\mathbf{x})$ converge toward a reasonable approximation of $h(\mathbf{x})$ and have similar distribution that satisfy:
\[
    \begin{aligned}
        & \mathrm{Cov}\left(h(\mathbf{x}), g^{bit}_{\mathbf{w}}(\mathbf{x})\right) = 0, \quad \mathrm{sign}(\mathbb{E}_{\mathbf{x}} \left[h(\mathbf{x}) \right]) > 0, \\
        & h(\mathbf{x}) \cdot g^{T+bit}_\mathbf{w}(\mathbf{x}) \sim h(\mathbf{x}) \cdot g^T_\mathbf{w}(\mathbf{x}),\\
    \end{aligned}\footnote{In the context of our experiments, the spike density produced by bit-plane encoding scales proportionally with the input magnitude, exhibiting behavior similar to the rate-coding distribution $g^{T}_{\mathbf{w}}(\mathbf{x})$. Under the assumption of a low timestep budget, these two distributions become closely aligned. Consequently, we adopt the approximation $g^{T}_{\mathbf{w}}(\mathbf{x}) \sim g^{bit}_{\mathbf{w}}(\mathbf{x})$, which motivates the initial assumption used in our analysis.}
\] 
there exist:
\[
    \mathrm{sup}\left(SNR\left(\delta^{bit}(\mathbf{x})\right)\right) \geq  \mathrm{sup}\left(SNR\left(\delta(\mathbf{x})\right)\right).
\] 
where:
\[
    \mathrm{sup}\left(SNR(\delta^{bit}(\mathbf{x}))\right) = \frac{\left\| \mathbb{E}_{\mathbf{x}} \left[ \delta^{bit}(\mathbf{x}) \right] \right\|^2_\infty}{\mathbb{E}_{\mathbf{x}}\left\|\delta^{bit}(\mathbf{x})-\mathbb{E}_{\mathbf{x}}\left[\delta^{bit}(\mathbf{x})\right]\right\|^2_\infty},
\]
\[
    \mathrm{sup}\left(SNR(\delta(\mathbf{x}))\right) = \frac{\left\| \mathbb{E}_{\mathbf{x}} \left[ \delta(\mathbf{x}) \right] \right\|^2_\infty}{\mathbb{E}_{\mathbf{x}}\left\|\delta(\mathbf{x})-\mathbb{E}_{\mathbf{x}}\left[\delta(\mathbf{x})\right]\right\|^2_\infty}
\] 
\end{theorem}
\begin{proof}
In order to prove Theorem \ref{main_theorem}, we first need some auxiliary lemmas to be proven. 
\begin{lemma} \label{lemma_norm}
Assuming bit plane concatenated SNN model $p^{bit}_{\mathbf{w}}(\mathbf{x})$ reasonably approximate $h(\mathbf{x})$ such that bit-plane Jacobian $g^{bit}_{\mathbf{w}}(\mathbf{x})$ satisfy:
\[
    \mathrm{Cov}\left(h(\mathbf{x}), g^{bit}_{\mathbf{w}}(\mathbf{x})\right) = 0, \quad \mathrm{sign}(\mathbb{E}_{\mathbf{x}} \left[h(\mathbf{x}) \right]) > 0,
\]
while bit plane gradient does not explode nor vanish over stochastic estimation, bounded as:
\begin{equation}
    G(\mathbf{w})^2 > \left\|  \mathbb{E}_{\mathbf{x}}\left[g^{bit}_\mathbf{w}(\mathbf{x})\right]\right\|_\infty > 0
    \label{eq:init_condition}
\end{equation}
for some scalar $G(\mathbf{w})$, there exist:
\[
    \left\| \mathbb{E}_{\mathbf{x}} \left[ \delta^{bit}(\mathbf{x}) \right] \right\|^2_\infty \geq \left\| \mathbb{E}_{\mathbf{x}} \left[ \delta(\mathbf{x}) \right] \right\|^2_\infty.
\]
\end{lemma}
\begin{proof}
In order for the resulted inequality in our lemma exist, we need to prove that under given condition, there exist:
\[
    \begin{cases}
        &\mathrm{sup}\left( \mathbb{E}_{\mathbf{x}} \left[ \delta^{bit}(\mathbf{x}) \right]\right) \geq \mathrm{sup}\left( \mathbb{E}_{\mathbf{x}} \left[ \delta(\mathbf{x}) \right] \right),\\
        &\mathrm{inf}\left( \mathbb{E}_{\mathbf{x}} \left[ \delta^{bit}(\mathbf{x}) \right]\right) \leq \mathrm{inf}\left( \mathbb{E}_{\mathbf{x}} \left[ \delta(\mathbf{x}) \right] \right).\\
    \end{cases}
\]
As noted from initial condition in Equation \ref{eq:init_condition}, we have:
\[
    \begin{aligned}
        & G(\mathbf{w})^2 > \left\|  \mathbb{E}_{\mathbf{x}}\left[g^{bit}_\mathbf{w}(\mathbf{x})\right]\right\|_\infty > 0 \\
        & \Leftrightarrow \mathrm{sup}\left( \mathbb{E}_{\mathbf{x}} \left[ g^{bit}_\mathbf{w}(\mathbf{x})\right]\right) \geq 0 \\
         & \Leftrightarrow \mathrm{sup}\left( \mathbb{E}_{\mathbf{x}} \left[h(\mathbf{x}) \right] \right) \cdot \mathrm{sup}\left( \mathbb{E}_{\mathbf{x}} \left[ g^{bit}_\mathbf{w}(\mathbf{x})\right]\right) \geq 0 \\
         & \Leftrightarrow \mathrm{sup}\left( \mathbb{E}_{\mathbf{x}} \left[h(\mathbf{x}) \right] \cdot \mathbb{E}_{\mathbf{x}} \left[ g^{bit}_\mathbf{w}(\mathbf{x})\right] + Cov\left(h(\mathbf{x}), g^{bit}_{\mathbf{w}}(\mathbf{x})\right)\right)\geq 0 \\
        & \Leftrightarrow \mathrm{sup}\left( \mathbb{E}_{\mathbf{x}} \left[h(\mathbf{x}) \cdot g^{bit}_\mathbf{w}(\mathbf{x})\right]\right) \geq 0 \\
        & \begin{aligned}
            &\Leftrightarrow \mathrm{sup}\left( \mathbb{E}_{\mathbf{x}} \left[h(\mathbf{x}) \cdot g^{T}_\mathbf{w}(\mathbf{x})\right]\right)+ \mathrm{sup}\left( \mathbb{E}_{\mathbf{x}} \left[h(\mathbf{x}) \cdot g^{bit}_\mathbf{w}(\mathbf{x})\right]\right)\\
            & \quad \geq \mathrm{sup}\left( \mathbb{E}_{\mathbf{x}} \left[h(\mathbf{x}) \cdot g^T_\mathbf{w}(\mathbf{x}) \right]\right)\\
        \end{aligned}\\
    \end{aligned}
\]
Under the property of expectation's sum and Minkowski's sum \cite{zakon2004mathematical}:
\[
    \begin{aligned}
        & \Leftrightarrow \mathrm{sup}\left( \mathbb{E}_{\mathbf{x}} \left[h(\mathbf{x}) \cdot g^{T+bit}_\mathbf{w}(\mathbf{x})\right]\right)
        \geq \mathrm{sup}\left( \mathbb{E}_{\mathbf{x}} \left[h(\mathbf{x}) \cdot g^T_\mathbf{w}(\mathbf{x}) \right]\right)\\
        & \Leftrightarrow \mathrm{sup}\left(\mathbb{E}_{\mathbf{x}} \left[\delta^{bit}(\mathbf{x})\right]\right) \geq \mathrm{sup}\left(\mathbb{E}_{\mathbf{x}} \left[\delta(\mathbf{x})\right]\right).\\
    \end{aligned}
\]
We can then use a similar process with infimum inequality of initial condition:
\[
    \begin{aligned}
        & G(\mathbf{w})^2 > \left\|  \mathbb{E}_{\mathbf{x}}\left[g^{bit}_\mathbf{w}(\mathbf{x})\right]\right\|_\infty > 0 \\
        & \Leftrightarrow \mathrm{inf}\left( \mathbb{E}_{\mathbf{x}} \left[ g^{bit}_\mathbf{w}(\mathbf{x})\right]\right) \leq 0 \\
        & \Leftrightarrow \mathrm{inf}\left( \mathbb{E}_{\mathbf{x}} \left[h(\mathbf{x}) \right] \right) \cdot \mathrm{inf}\left( \mathbb{E}_{\mathbf{x}} \left[ g^{bit}_\mathbf{w}(\mathbf{x})\right]\right) \leq 0 \\
         & \Leftrightarrow \mathrm{inf}\left( \mathbb{E}_{\mathbf{x}} \left[h(\mathbf{x}) \right] \cdot \mathbb{E}_{\mathbf{x}} \left[ g^{bit}_\mathbf{w}(\mathbf{x})\right] + Cov\left(h(\mathbf{x}), g^{bit}_{\mathbf{w}}(\mathbf{x})\right) \right) \leq 0 \\
        & \Leftrightarrow \mathrm{inf}\left( \mathbb{E}_{\mathbf{x}} \left[h(\mathbf{x}) \cdot g^{bit}_\mathbf{w}(\mathbf{x})\right]\right) \leq 0 \\
        & \begin{aligned}
            &\Leftrightarrow \mathrm{inf}\left( \mathbb{E}_{\mathbf{x}} \left[h(\mathbf{x}) \cdot g^{T}_\mathbf{w}(\mathbf{x})\right]\right)+ \mathrm{inf}\left( \mathbb{E}_{\mathbf{x}} \left[h(\mathbf{x}) \cdot g^{bit}_\mathbf{w}(\mathbf{x})\right]\right)\\
            & \quad \leq \mathrm{inf}\left( \mathbb{E}_{\mathbf{x}} \left[h(\mathbf{x}) \cdot g^T_\mathbf{w}(\mathbf{x}) \right]\right)\\
        \end{aligned}\\
        & \Leftrightarrow \mathrm{inf}\left( \mathbb{E}_{\mathbf{x}} \left[h(\mathbf{x}) \cdot g^{T+bit}_\mathbf{w}(\mathbf{x})\right]\right)
        \leq \mathrm{inf}\left( \mathbb{E}_{\mathbf{x}} \left[h(\mathbf{x}) \cdot g^T_\mathbf{w}(\mathbf{x}) \right]\right)\\
        & \Leftrightarrow \mathrm{inf}\left(\mathbb{E}_{\mathbf{x}} \left[\delta^{bit}(\mathbf{x})\right]\right) \leq \mathrm{inf}\left(\mathbb{E}_{\mathbf{x}} \left[\delta(\mathbf{x})\right]\right)\\
    \end{aligned}
\]
From here we can conclude that there exist:
\[
    \left\| \mathbb{E}_{\mathbf{x}} \left[ \delta^{bit}(\mathbf{x}) \right] \right\|^2_\infty \geq \left\| \mathbb{E}_{\mathbf{x}} \left[ \delta(\mathbf{x}) \right] \right\|^2_\infty
\]
under given condition.
\end{proof}
\begin{lemma} \label{lemma_var}
Assuming both gradient of baseline SNN $g_\mathbf{w}(\mathbf{x})$ and bit plane code concatenated SNN $g^{bit}_\mathbf{w}(\mathbf{x})$ converge toward a reasonable approximation of $h(\mathbf{x})$ and have similar distribution that satisfy:
\[
    h(\mathbf{x}) \cdot g^{T+bit}_\mathbf{w}(\mathbf{x}) \sim h(\mathbf{x}) \cdot g^{T}_\mathbf{w}(\mathbf{x}),
\]
there exist:
\[
    \mathbb{E}_{\mathbf{x}}\left\|\delta^{bit}(\mathbf{x})-\mathbb{E}_{\mathbf{x}}\left[\delta^{bit}(\mathbf{x})\right]\right\|^2_\infty = \mathbb{E}_{\mathbf{x}}\left\|\delta(\mathbf{x})-\mathbb{E}_{\mathbf{x}}\left[\delta(\mathbf{x})\right]\right\|^2_\infty.
\] 
\end{lemma}
\begin{proof}
Since we have the initial condition that:
\[
    h(\mathbf{x}) \cdot g^{T+bit}_\mathbf{w}(\mathbf{x}) \sim h(\mathbf{x}) \cdot g^{T}_\mathbf{w}(\mathbf{x}),
\] 
then:
\begin{equation}
    \begin{aligned}
        & g^{T+bit}_\mathbf{w}(\mathbf{x}) \sim g^{T}_\mathbf{w}(\mathbf{x})\\ 
        & \Leftrightarrow Var\left(\delta^{bit}(\mathbf{x})\right) = Var\left(\delta(\mathbf{x}) \right) \\
        & \Leftrightarrow \left\|\delta^{bit}(\mathbf{x})-\mathbb{E}_{\mathbf{x}}\left[\delta^{bit}(\mathbf{x})\right]\right\|^2_\infty =  \left\|\delta(\mathbf{x})-\mathbb{E}_{\mathbf{x}}\left[\delta(\mathbf{x})\right]\right\|^2_\infty\\
        & \Leftrightarrow \mathbb{E}_{\mathbf{x}}\left\|\delta^{bit}(\mathbf{x})-\mathbb{E}_{\mathbf{x}}\left[\delta^{bit}(\mathbf{x})\right]\right\|^2_\infty = \mathbb{E}_{\mathbf{x}}\left\|\delta(\mathbf{x})-\mathbb{E}_{\mathbf{x}}\left[\delta(\mathbf{x})\right]\right\|^2_\infty.\\
    \end{aligned}
\end{equation}
\end{proof}
From results of Lemma \ref{lemma_norm} and \ref{lemma_var}, we can conclude that there exist:
\[
    \begin{aligned}
        & \frac{\left\| \mathbb{E}_{\mathbf{x}} \left[ \delta^{bit}(\mathbf{x}) \right] \right\|^2_\infty}{\mathbb{E}_{\mathbf{x}}\left\|\delta^{bit}(\mathbf{x})-\mathbb{E}_{\mathbf{x}}\left[\delta^{bit}(\mathbf{x})\right]\right\|^2_\infty} \geq
        \frac{\left\| \mathbb{E}_{\mathbf{x}} \left[ \delta(\mathbf{x}) \right] \right\|^2_\infty}{\mathbb{E}_{\mathbf{x}}\left\|\delta(\mathbf{x})-\mathbb{E}_{\mathbf{x}}\left[\delta(\mathbf{x})\right]\right\|^2_\infty}\\
        & \Leftrightarrow \mathrm{sup}\left(SNR\left(\delta^{bit}(\mathbf{x})\right)\right) \geq  \mathrm{sup}\left(SNR\left(\delta(\mathbf{x})\right)\right).\\
    \end{aligned}
\] 
\end{proof}

On the other hand, there are special situation where the $L_2$ norm $\left\| \delta(\mathbf{x}) \right\|_2$ directly scale with model Jacobian $g_\mathbf{w}(\mathbf{x})$ dimensions. If target function $h(\mathbf{x})$ is zero mean normally distributed $\mathcal{N}(0, \sigma^2_h)$, $g_\mathbf{w}(\mathbf{x})$ have sufficiently small variance and does not saturate, $\delta(\mathbf{x})$ would follow a zero mean normal distribution (denoted $\mathcal{N}(0, \sigma^2_\delta)$) and its norm can be directly decomposed to:
\[
\left\| \delta(\mathbf{x}) \right\|_2 = \sqrt{\sum_{i=1}^{n_1^\delta} \sum_{j=1}^{n_2^\delta} \delta_{ij}^2}.
\]
where each squared entry \( \delta_{ij}^2 \) would follow a chi-squared distribution with 1 degree of freedom with $\delta_{ij}^2 \sim  \sigma^2_\delta\chi^2(1)$, the sum of all squared entries follows a chi-squared distribution with \( n_1^\delta n_2^\delta \) degrees of freedom (by Cochran's theorem\cite{cochran1934distribution}):
\[
\sum_{i=1}^{n_1^\delta} \sum_{j=1}^{n_2^\delta} g_{ij}^2 \sim  \sigma^2_\delta\chi^2(n_1^\delta n_2^\delta).
\]
From the expectation property of the chi-squared distribution where $\chi^2(n_1^\delta n_2^\delta ) \sim Gamma(\alpha=nk/2, \theta=2/n)$ of $n$ samples with $k$ degrees of freedom, we have:
\[
E\left[  \sigma^2_\delta\chi^2(n_1^\delta n_2^\delta ) \right] =  \sigma^2_\delta n_1^\delta n_2^\delta .
\]
Taking the square root gives:
\[
\left\| \delta({\mathbf{x}}) \right\|_2 \approx \left| \sigma_\delta \right| \sqrt{ n_1^\delta n_2^\delta}.
\]
Thus, $\left\| \delta({\mathbf{x}}) \right\|_2$ grows asymptotically as \( O(\left| \sigma_\delta \right| \sqrt{ n_1^\delta n_2^\delta}) \). We can also quantify the probability for lower bound and upper bound of $\left\| \delta({\mathbf{x}}) \right\|$ using the consequences of the standard Laurent-Massart bounds\cite{laurent2000adaptive}:
\begin{equation}
    \begin{aligned}
        & k = n_1^\delta n_2^\delta\\
        & \begin{aligned}
            A := \left\| \delta({\mathbf{x}}) \right\|_2 \in & [\sigma^2_\delta k - 2\sigma^2_\delta k^{\frac{1}{2+\alpha}},\\
            & \sigma^2_\delta k + 2 \sigma^2_\delta k^{\frac{1}{2+\alpha}} + 2 \sigma^2_\delta k^{\alpha}]\\
        \end{aligned}\\
        & Pr\left(A\right) \ge 1 - e^{-k^{\alpha}} \left(\forall \alpha \in \mathbb{R}\right)\\
    \end{aligned}
\end{equation}
It is clear that as $k$ increase, lower bound, upper bound of $\left\| \delta({\mathbf{x}}) \right\|_2$ and probability of $\left\| \delta({\mathbf{x}}) \right\|_2$ is within the established range also increase.
\end{appendices}

\begin{IEEEbiography}[{\includegraphics[width=1in, height=1.25in, clip, keepaspectratio]{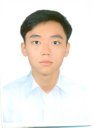}}]{Luu Trong Nhan} (Member, IEEE) received the B.Sc. degree and M.Sc. degree in Computer Science from The University of Aizu, Japan, in 2024 and 2025 respectively. He is currently working as Research Assistant at College of Information and Communication Technology, Can Tho University, Vietnam. His research interests include machine learning, artificial intelligent algorithm, computational neuroscience and quantum computing.
\end{IEEEbiography}

\begin{IEEEbiography}[{\includegraphics[width=1in,height=1.25in,clip,keepaspectratio]{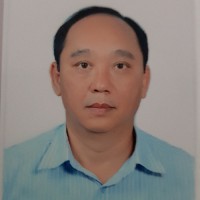}}]{Luu Trung Duong} received the B.E. degree in Informatics from Can Tho University (CTU), Vietnam, in 1997 and M.Sc. degree in Mobile and Distributed Computer Network from Leeds Beckett University, England, in 2003. He has been working at Can Tho University since 1997 to present, starting as a Network Manager and currently is the Director of Center for Digital Transformation and Communication. His research interests include large scale networking solution design, computer security and network management.
\end{IEEEbiography}

\begin{IEEEbiography}[{\includegraphics[width=1in,height=1.25in,clip,keepaspectratio]{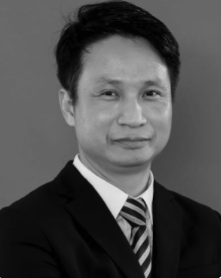}}]{Pham Ngoc Nam} (Member, IEEE) received the bachelor’s degree in electronics engineering from the Hanoi University of Science and Technology (HUST), in 1997, and the M.S. degree in artificial intelligence and the Ph.D. degree in electrical engineering from KU Leuven, Belgium, in 1999 and 2004, respectively. He is currently the Vice Dean of the College of Engineering and Computer Science, VinUniversity, Vingroup. He is also a Visiting Scholar with Cornell University. He has been the PI of one key national project and three ministerial-level projects. He is the author or coauthor of 100 scientific articles, including more than 30 ISI and Scopus publications. His research interests include artificial intelligence, QoS/QoE management for multimedia applications, reconfigurable computing, and low-power embedded system design. He has been a key member of four other national projects.
\end{IEEEbiography}

\begin{IEEEbiography}[{\includegraphics[width=1in,height=1.25in,clip,keepaspectratio]{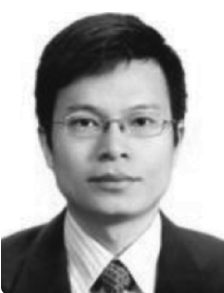}}]{Truong Cong Thang} (Senior Member, IEEE) received the B.E. degree from Hanoi University of Technology, Vietnam, in 1997 and Ph.D. degree from KAIST, Korea, in 2006. From 1997 to 2000, he worked as an engineer in Vietnam Post \& Telecom (VNPT). From 2006 to 2011, he was a Member of Research Staff at Electronics and Telecommunications Research Institute (ETRI), Korea. He was also an active member of Korean delegation to standard meetings of ISO/IEC and ITU-T from 2002 to 2011. Since 2011, he has been an Associate Professor of University of Aizu, Japan. His research interests include multimedia networking, image/video processing, content adaptation, IPTV and MPEG/ITU standards.
\end{IEEEbiography}

\vfill
\EOD
\end{document}